\pdfoutput=1

\documentclass[11pt]{article}

\usepackage{EMNLP2022}
\usepackage{times}
\usepackage{latexsym}
\usepackage[T1]{fontenc}

\usepackage[utf8]{inputenc}
\usepackage{microtype}
\usepackage{marvosym}
\usepackage{multirow}
\usepackage{amsthm}
\usepackage{algorithm}
\usepackage{algorithmic}
\usepackage{amsmath}
\usepackage{booktabs}
\usepackage{latexsym}
\usepackage{amssymb}
\usepackage{arydshln}
\usepackage{inconsolata}
\usepackage{graphicx}
\usepackage{caption}
\usepackage{subcaption}

\newtheorem{remark}{\textbf{Remark}}

\def \mL{{\mathcal L}}

\def \v{{\bf v}}

\def \t{{\bf t}}

\def \y{{\bf y}}
\def \p{{\bf p}}
\def \V{{\bf V}}
\def \tV{\tilde{{\bf V}}}

\def \mR{\mathbb{R}}

\title{LiteVL: Efficient Video-Language Learning with Enhanced Spatial-Temporal Modeling}

\author{Dongsheng Chen$^1$, Chaofan Tao$^2$, Lu Hou$^3$, Lifeng Shang$^3$, Xin Jiang$^3$, Qun Liu$^3$ \\
  $^1$Peking University\quad$^2$The University of Hong Kong \quad $^3$Huawei Noah's Ark Lab\\
\texttt{chends@stu.pku.edu.cn, cftao@connect.hku.hk} \\
  \texttt{\{houlu3,shang.lifeng,jiang.xin,qun.liu\}@huawei.com}}

\begin{document}
\maketitle
\begin{abstract}
Recent large-scale video-language pre-trained models
have shown appealing performance on various downstream tasks.
However, the pre-training process is computationally expensive due to the requirement of millions of video-text pairs and the redundant data structure of each video.
To mitigate these problems, we propose LiteVL, 
which adapts a pre-trained image-language model BLIP into a video-text model directly on downstream tasks,
without heavy pre-training.
To enhance the temporal modeling lacking in the image-language model,
we propose to add temporal attention modules in the image encoder of BLIP with dynamic temporal scaling. Besides the model-wise adaptation, we also propose a non-parametric pooling mechanism to adaptively reweight the fine-grained video embedding conditioned on the text.
Experimental results on text-video retrieval and video question answering show that the proposed LiteVL even outperforms previous video-language pre-trained models by a clear margin, though without any video-language pre-training.

\end{abstract}

\section{Introduction}
\label{sec:intro}
The increasing popularity of videos in various social media has aroused the interest in efficient modeling of videos and their connections with other modalities like texts. 
Video-language modeling targets at learning a shared 
 multimodal semantic space for videos and texts to facilitate downstream tasks like text-video retrieval and video question answering (VideoQA). 
Previous video-language modeling
usually relies on  pre-training on a large-scale video-text pair, via
video-text contrastive learning \citep{DBLP:journals/corr/abs-2104-08860,li2021align,gorti2022xpool}, video-text matching \citep{DBLP:conf/cvpr/FanZZW0H19,luo2020univl,li2021align}, masked language modeling \citep{DBLP:conf/naacl/DevlinCLT19} and masked frame modeling \citep{DBLP:conf/iccv/SunMV0S19};
or extra 
object detectors to extract
fine-grained visual features \citep{zhu2020actbert,chen2020uniter}. 
However, both pre-training on video-text pairs and using
an off-the-shelf detector are  computationally expensive and inefficient. Inaccurate detection results on limited categories may also lead to inferior performance.

In the unimodal video domain, 
TimeSformer \citep{timesformer} fails to pre-train video encoder directly on large-scale video datasets, but obtains good performance by initializing from a pre-trained Transformer-based ViT~\citep{vit} image encoder and training additional temporal attention modules directly on downstream video tasks. 
Similarly, ViViT~\citep{vivit} also takes advantage of the already well-learned spatial visual representation in a pre-trained ViT model, and effectively adapts it for video tasks by directly fine-tuning on comparatively small downstream video datasets.

Inspired by TimeSformer and ViViT,
in this paper, we also consider extending an image-language pre-trained model for video-text tasks without  pre-training on video-text pairs.
This requires us to not only leverage the already-learned alignment between spatial visual information and text in the image-language models, 
but also capture the additional temporal dependency efficiently.
Thus we propose a simple yet efficient video-language model LiteVL, initialized from a recent pre-trained image-language model BLIP, but with both model-wise and feature-wise enhancement of temporal information.
For model-wise enhancement, we propose to explicitly insert temporal attention layers with learnable scalings into the original image backbone, which can be adjusted for each downstream task.
For feature-wise enhancement, we design a non-parametric pooling method to learn fine-grained spatial-temporal video features conditioned on the text description.

Empirical results on various  tasks demonstrate that the proposed model, LiteVL, outperforms previous state-of-the-art methods by a clear margin, even without any video-language pre-training or the usage of object detectors. 
In particular,  LiteVL achieves 50.8\% R1 score on \textbf{MSRVTT}-9k dataset in the task of  text-video retrieval, and 42.9\% accuracy on \textbf{MSRVTT-QA} dataset in the task of video question answering.  
Visualizations also demonstrate that our LiteVL captures important spatial-temporal information with fine-grained video-text alignment. 


\begin{figure*}[t!]
    \centering
    \begin{subfigure}[b]{0.76\textwidth}
        \centering
  \includegraphics[width=1\textwidth]{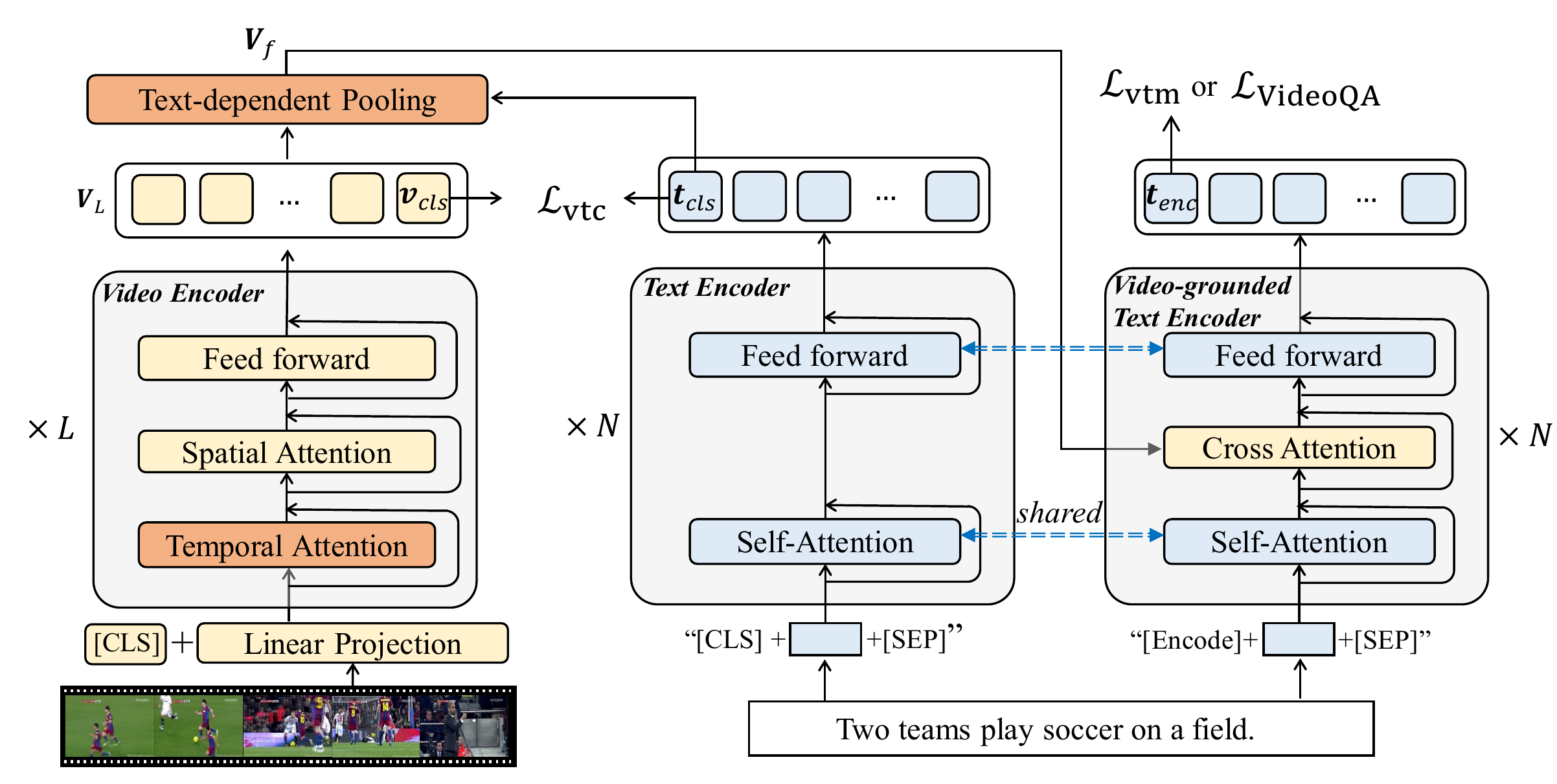}
        \caption{Architecture of LiteVL.}
        \label{fig:model_structure}
    \end{subfigure}
    \hfill
    \begin{subfigure}[b]{0.23\textwidth}
        \centering
        \includegraphics[width=\textwidth]{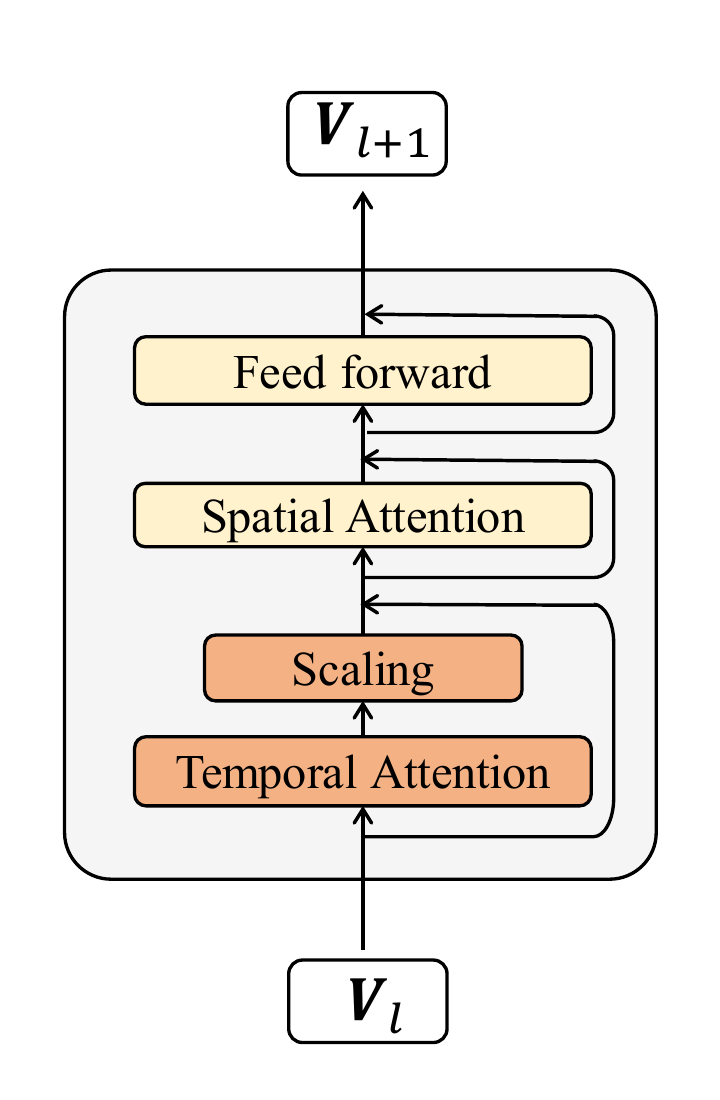}
        \caption{Temporal Scaling.}
        \label{fig:scaling}
    \end{subfigure}
      \caption{(a) The architecture of LiteVL. The model is initialized from the pre-trained image-language model BLIP, but is equipped with additional temporal attention modules and text-dependent pooling, to quickly adapt to video-language downstream tasks without pre-training.
    (b) The proposed dynamic temporal scaling, 
    which adjusts the scale of the newly-added temporal attention 
    according to each downstream task.
      }
    \label{fig:model}
\end{figure*}

\section{Related Work}
\label{sec:related_work}
\subsection{Vision Transformers}
Transformers \citep{DBLP:conf/nips/VaswaniSPUJGKP17}, originally designed for natural language tasks, have recently been applied to the computer vision domain to model images and videos
\citep{vit,DBLP:conf/iccv/LiuL00W0LG21,DBLP:conf/icml/TouvronCDMSJ21,DBLP:conf/iccv/WangX0FSLL0021,DBLP:conf/nips/HanXWGXW21}.  
ViT \citep{vit} is one of the most representative vision transformers, which
processes each image as a sequence of image patches, and
achieves remarkable performances on various image tasks.

Compared with image tasks, video understanding is more challenging because the additional temporal dimension brings a more complicated
spatial-temporal information.  
To model the intertwined dependency of the spatial and temporal dimensions efficiently, video Transformers
TimeSformer \citep{timesformer} and ViViT \citep{vivit} use the parameters of a well-trained image Transformer for initialization, and further
design different variants of spatial-temporal attention mechanisms to capture the spatial-temporal dependencies. 


\subsection{Video-Language Modeling}
\label{sec:relatedwork_vlmodel}
The core of video-language  models lies in
modeling the interaction between the two modalities.
Depending on whether using video-text pairs for pre-training,
existing video-language models can be divided into two categories.

The main branch of works explicitly designs the spatial-temporal structure in video encoders, and pre-train with abundant video-text pairs to directly align videos and texts.
Among these works, ALPRO~\citep{li2021align}, Frozen~\citep{DBLP:conf/iccv/BainNVZ21}, and BridgeFormer~\citep{DBLP:journals/corr/abs-2201-04850} use WebVid2M \citep{DBLP:conf/iccv/BainNVZ21}  which contains 2.5M video-text pairs collected from the web for pre-training. 
Image-text pairs like CC3M \citep{sharma2018conceptual} and VG \citep{krishnavisualgenome} are also often used to enhance
the spatial information in this alignment. NoiseEST~\citep{DBLP:conf/aaai/AmraniBRB21}, VideoCLIP~\citep{DBLP:conf/emnlp/XuG0OAMZF21}, and CLIP4Clip~\citep{DBLP:journals/corr/abs-2104-08860} pretrain the model with a large-scale dataset HowTo100M \citep{DBLP:conf/cvpr/FanZZW0H19} which contains 136M video-text pairs.
The other branch of works does not rely on video-text pre-training.
Instead, they extend a pre-trained image-text model to extract video features, and directly fine-tune on downstream tasks. In ClipBERT~\citep{DBLP:conf/cvpr/LeiLZGBB021}, BLIP~\citep{blip}, and X-Pool~\citep{gorti2022xpool},  each video is viewed as a collection of images, whose representations obtained from an image encoder are then used to represent the video for later interaction with the text. 

\section{Method}
\label{sec:method}
In this section, we propose a method to efficiently extend an image-language
pre-trained  model to a video-language model, without pre-training on video-text pairs or the use of object detectors.

In Section~\ref{sec:architecture},
we first introduce 
the model architecture. Then we propose to enhance the temporal information in both model-wise and feature-wise manners.
For model-wise enhancement, we propose to insert temporal attention layers  with learnable scalings into the original image backbone (Section~\ref{sec:scaling}).
For feature-wise enhancement, we design a non-parametric pooling method to learn fine-grained spatial-temporal video features conditioned on the text description (Section~\ref{sec:pooling}).

\subsection{Model Architecture}
\label{sec:architecture}

As is shown in the Figure \ref{fig:model}, our framework contains three parts: a video encoder,
a text encoder,
and a video-grounded text encoder.
We initialize our framework based on the recently proposed image-language model BLIP
\citep{blip} trained over massive image-text pairs.

\paragraph{Video Encoder.}
To enhance the temporal dependency of the video encoder,
following TimeSformer~\citep{timesformer}, we insert additional temporal attention modules into the original BLIP image encoder, whose weights are initialized with the original spatial attention modules (Figure~\ref{fig:model_structure}).
We use the Divided Space-Time Attention
proposed in TimeSformer. 
We first compute the temporal attention by comparing each patch with the patches at the same spatial location in different frames, and then compute the spatial attention in each frame separately.

\paragraph{Text Encoder.}
The text encoder is a BERT model, initialized with the original text encoder in BLIP.
A $\texttt{[CLS]}$ token is appended to the beginning of the text sequence. 
This unimodal text encoder uses bi-directional self-attention and uses the output embedding of the $\texttt{[CLS]}$ token to summarize the text sequence.

\paragraph{Video-grounded Text Encoder.}
This encoder shares the parameters with the unimodal text encoder.
Moreover, to fuse the video features from the video encoder, 
one additional cross-attention layer is added between the self-attention layer and the feed-forward network for each transformer layer. 
Following BLIP~\citep{blip}, we also use a special $\texttt{[Encode]}$ token before the text sequence, and its output embedding is used as the multimodal representation of the video-text pair.

 \subsection{Dynamic Temporal Scaling}
 \label{sec:scaling}

We wish to preserve the spatial representation and its alignment with the text encoder learned by the image-language pre-trained model, as well as learn temporal expressivity for video-language tasks.  As will be shown in Table~\ref{tab:ablation_scaling_pooling}, 
directly using TimeSformer
yields better results than the original ViT.
To provide more sufficient temporal expressiveness of the video encoder,
we propose to learn a set of scalings that dynamically adjust the newly inserted temporal attention modules according to each specific task, as shown in Figure~\ref{fig:scaling}.

Specifically, denote the output feature 
after the  temporal attention at the $l$-th Transformer layer as $\V_l^{\text{TAttn}} \in \mR^{S \times T \times D}$. For $t$-th frame $[\V_l^{\text{TAttn}}]_t \in \mR^{S \times D}$, we add a learnable scaling factor  $\alpha_{l,t} \in \mR$ with a tanh-gating mechanism as:
\begin{equation}
    \alpha_{l,t} =  \text{tanh}(\gamma_{l,t})+1,
    \label{eq:scaling}
\end{equation}
where $\gamma_{l,t}$ is a learnable scalar initialized at $0$. 
Then the scaled output of temporal attention $\V_l^{\text{TAttn}}$ is calculated as:
\begin{equation}
[\V_l^{\text{TAttn}}]_t = \alpha_{l,t} \cdot [\V_l^{\text{TAttn}}]_t,
\label{equ:alt}
\end{equation}
before the residual connection. Note the $\texttt{[CLS]}$ token is kept but not involved in the computation of scaling.
The choice of tanh-gating ensures that $\alpha_{l,t}$ ranges from 0 to 2. 
Initially, our model is equivalent to TimeSformer (i.e., $\alpha_{l,t}$=1, treat each frame equally), 
and then explicitly reweight the frames
in each transformer block during the fine-tuning stage.  
When  $\alpha_{l,t}$ reduces to 0, the video encoder degenerates to the ViT used in the original BLIP model, which does not consider any temporal dependency in extracting the video features.

\subsection{Text-dependent Pooling}

\label{sec:pooling}
Before interacting with the textual modality via cross-attention or self-attention, 
previous methods  directly concatenate the features from all frames with equal importance~\citep{blip,DBLP:journals/corr/abs-2201-04850},
or
aggregate the video features with heuristic mean/max pooling  methods spatially or temporally~\citep{DBLP:journals/corr/abs-2104-08860,li2021align}.
However, not all frames or spatial positions are equally representative for the whole video, and different frames or positions have different semantic similarities to the textual query (e.g., textual description in text-video retrieval tasks or textual question in video question answering tasks. For example, given a video description  ``\texttt{a golf player is trying to hit the ball into the pit}'', the video encoder is expected to focus on the object of interest (i.e., ball) and the motion of hitting across the frames.

\begin{figure}
    \centering
    \includegraphics[width=0.9\linewidth]{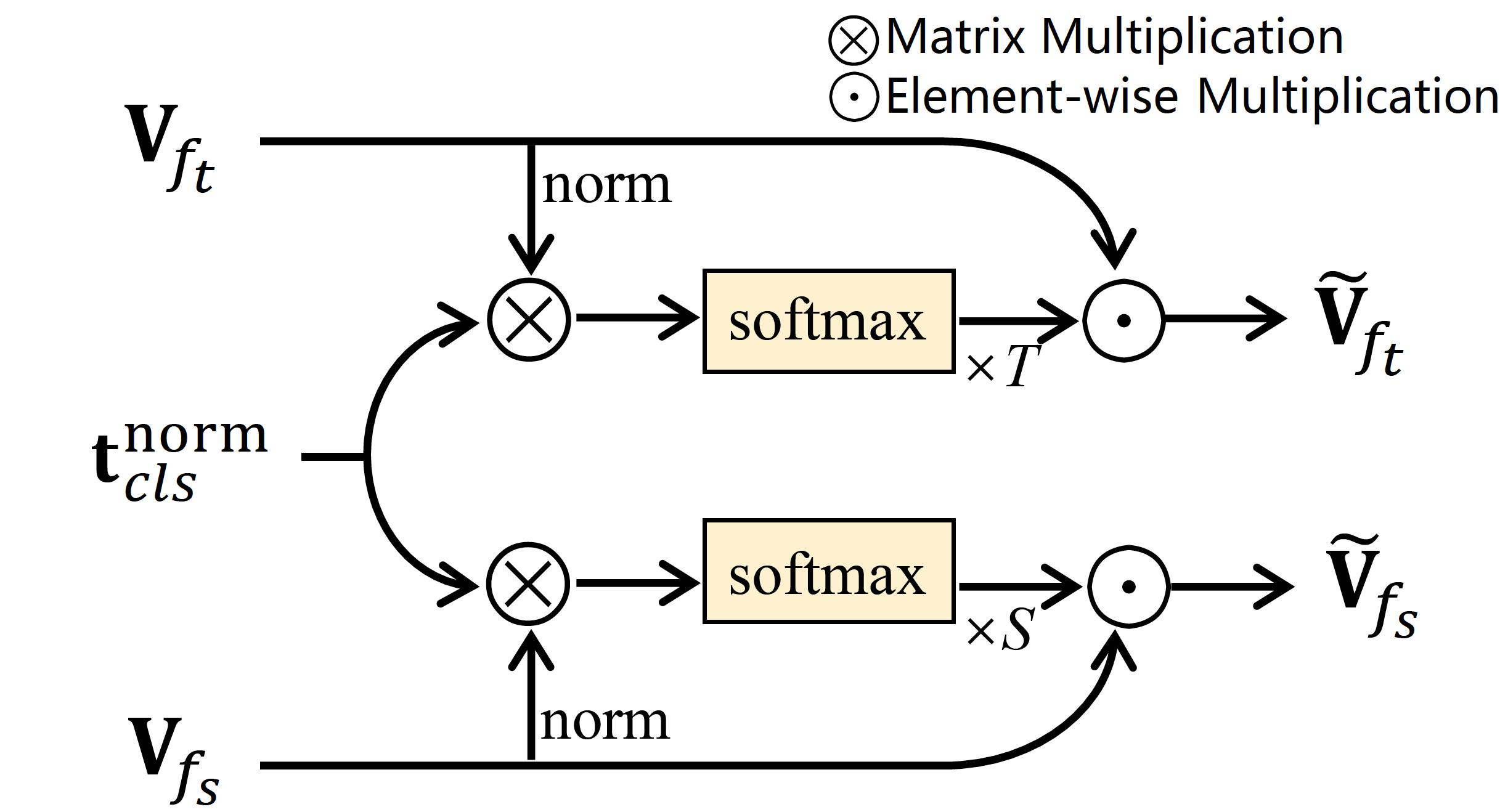}
    \caption{Illustration of text-dependent pooling.  {We reweight the pooled spatial and temporal video features based on the similarities between the normalized text feature $\t_{cls}^\text{norm}$ and visual features $\V_{f_t}$ and  $\V_{f_s}$. }}
	\label{fig:pool}
\end{figure}

As illustrated in Figure \ref{fig:pool}, we design 
a non-parametric text-dependent pooling
to reweight the video features spatially and temporally depending on the corresponding textual query, enabling fine-grained video-text alignment.

Specifically, given a video with $T$ frames,
each frame is pachified into $S$ patches, 
and a $\texttt{[CLS]}$ token is inserted before the $ST$ patches. 
Denote the original output embedding of the video encoder as $\V_L\in \mathbb{R}^{ (1+ST) \times D}$.
$\V_{f_t} \in \mR^{T \times D}$ and $\V_{f_s} \in \mR^{S \times D}$ are the
video features  pooled by averaging $\V_L$ along the spatial and temporal dimension, respectively. 
Note that the feature of $\texttt{[CLS]}$ token is not involved in averaging. 
Denote  $\t_{cls} \in \mR^{ D} $ as the output embedding of the $\texttt{[CLS]}$ token obtained from the text encoder.

Intuitively, the more similar a visual feature is to the text description, the more representative it is for understanding the content of the whole video. Thus we compute the similarity between the $\ell_2$ normalized features of each frame in $\V_{f_t}^\text{norm}$ and the text feature $\t_{cls}^\text{norm}$, and reweight the features in $\V_{f_t}$ as:
\begin{eqnarray}
    {\bf g}_t &=&\text{softmax}(\V_{f_t}^\text{norm}\t_{cls}^\text{norm}/\tau) , \label{eq:gamma} \\
        \tV_{f_t} &=& T \cdot {\bf g}_t \odot \V_{f_t},  \nonumber
\end{eqnarray}
where $\odot$ means
element-wise multiplication, and  $\tau$ is the temperature which controls the sharpness of the weight distribution.
We multiply the weights from the softmax function by the
 number of frames $T$, such that the sum of  total weights keeps the same as direct concatenation.
Similarly, we compute the similarity between the $\ell_2$ normalized features of each spatial position in $\V_{f_s}^\text{norm}$ and the text feature $\t_{cls}^\text{norm}$, and reweight the features in $\V_{f_s}$ as:
\begin{equation}
    \tV_{f_s} = S \cdot \text{softmax}(\V_{f_s}^{\text{norm}}\t_{cls}^{\text{norm}}/\tau) \odot \V_{f_s}.
\end{equation}
The final aggregated video feature to be fed to the video-grounded text encoder is a concatenation of  $\tV_{f_t},\tV_{f_s}$, and the original video feature $\V_L$:
\begin{equation}
     \V_f = [\tV_{f_t},\tV_{f_s}, \V_L].
     \label{eq:concat_video_feature}
\end{equation}

\begin{remark}
\label{rmk:vanilla_pooling}
Besides using the text to reweight the aggregated features after spatial pooling (i.e., $\V_{f_t}$) and temporal pooling (i.e., $\V_{f_s}$),
one simple baseline is to directly concatenate them with the
original features $\V_L$ to compose $\V_f$ as:
    $
    \V_f = [\V_{f_t}, \V_{f_s}, \V_L] \in \mR^{(1+T+S+ST) \times D}.
    $
We dub it as \textbf{vanilla pooling}. 
Despite its simplicity, this  pooling
achieves competitive performance (Table \ref{tab:ablation_scaling_pooling}).
\end{remark}

\begin{table*}[!htbp]
\centering
\scalebox{0.82}{
	\begin{tabular}{l cccc  ccccc c c c c c }
		\toprule

        \multirow{2}{*}[-3pt]{\textbf{Methods}} &
         \multirow{2}{*}[-3pt]{\textbf{\shortstack{Pre-Training\\Data}}}&
		\multicolumn{4}{c}{\textbf{MSRVTT}-7k} &\multicolumn{4}{c}{\textbf{MSRVTT}-9k} & \multicolumn{4}{c}{\textbf{DiDeMo}} \\ 
		\cmidrule(lr){3-6} \cmidrule(lr){7-10}\cmidrule(lr){11-14}  &
		&
		\textbf{R1}$\uparrow$& \textbf{R5}$\uparrow$& \textbf{R10}$\uparrow$& \textbf{MdR}$\downarrow$
		&
		\textbf{R1}$\uparrow$& \textbf{R5}$\uparrow$& \textbf{R10}$\uparrow$& \textbf{MdR}$\downarrow$
	&\textbf{R1}$\uparrow$& \textbf{R5}$\uparrow$& \textbf{R10}$\uparrow$& \textbf{MdR}$\downarrow$ \\
		\midrule [\heavyrulewidth]
		BLIP$^\dagger$   & 14M &33.6& 55.5 & 64.8& 4 &33.6& 55.5 & 64.8& 4& 35.6 &58.4 & 66.4& 3 \\
		BLIP$^{384\dagger}$ &14M &30.8 & 52.2 & 62.2 & 5 &30.8 & 52.2 & 62.2 & 5& 31.6& 58.2& 65.8 &3 \\
		BLIP  & 14M &42.5 &68.2 &79.3 &2  &44.1&70.9&80.1&2&51.3 &78.1 &84.6 & 1   \\
		LiteVL$_S$  
		&-& \textbf{44.5} & \textbf{70.3} & \textbf{80.2} & \textbf{2} &\textbf{46.7}&\textbf{71.8} &\textbf{81.7}&\textbf{2}& \textbf{53.7}& \textbf{79.6}& \textbf{87.0}& \textbf{1}\\		
        \midrule\midrule
		BLIP$^\dagger$ & 129M  &  41.4&63.3 & 72.8 & 2 &  41.4&63.3 & 72.8 & 2&40.0 &63.1 & 72.4 & 2 \\
		BLIP$^{384\dagger}$  & 129M&  40.5&62.8 &71.4  & 2 &  40.5&62.8 &71.4  & 2&37.3 & 61.7& 69.4 & 3 \\ 
		BLIP$_\mathrm{coco}^\dagger$  &129M & 40.3 &63.6 &72.1  & 2 & 40.3 &63.6 &72.1  & 2&40.0 &64.6 & 73.1 & 2  \\
		BLIP$_\mathrm{coco}^\mathrm{384 \dagger}$ &129M &43.3  & 65.6& 74.7 & 2 &43.3  & 65.6& 74.7 & 2 &43.2 & 69.3&75.9  &2   \\
		BLIP & 129M &48.2 &74.3 & 82.9&  2& 49.7&74.7&83.8&2 &51.6& 79.8 & 86.8& 1  \\
		LiteVL$_L$  & -& 48.9 &74.5 & \textbf{83.6} &2 &50.8 &\textbf{76.3} &84.4 &1&\textbf{53.4} &\textbf{80.7} & 87.0 & 1 \\
	LiteVL$_{L \mathrm{coco}}^\mathrm{384}$  &- &\textbf{49.7} &\textbf{75.5} & 83.3  & \textbf{2}& \textbf{51.5} &75.9 &\textbf{85.7}  &\textbf{1} &53.2 &79.6 & \textbf{87.5} &\textbf{1}   \\
\midrule [\heavyrulewidth]

	\end{tabular}}\
			\vspace{-0.05in}

	\caption{Comparison of  LiteVL and BLIP on text-video retrieval tasks.
	The default resolution is 224$\times$224 per video frame, and the superscript ``384'' means increasing it to 384$\times$384. 
	The subscript ``$\mathrm{coco}$'' means training with an extra COCO retrieval dataset.
	$\dagger$ means zero-shot inference used in BLIP by default.
    }
    \label{tab:ablation_timesformer}
\end{table*}

\subsection{Training Objectives}
After obtaining the aggregated video features $\V_f$, 
we feed them to  each cross-attention layer of the video-grounded text encoder. 
Consider a training batch with $B$ video-text pairs.
For the $k$-th video-text pair, 
denote the $\ell_2$ normalized output embeddings
of the $\texttt{[CLS]}$ tokens from the video encoder and the  text encoder as $\v^k_{cls}$ and $\t^k_{cls}$\footnote{Following BLIP~\citep{blip}, 
before normalization, the $\texttt{[CLS]}$ tokens from the video and text encoders are mapped to a lower-dimensional space 
through separate linear projections. 
With a slight abuse of notation,
this $\t^k_{cls}$ has reduced dimension and is not the same as $\t_{cls}^{\text{norm}}$ in Eq.(\ref{eq:gamma}).},
respectively. 
The output embedding of the $\texttt{[Encode]}$ token of the video-grounded text encoder is denoted as $\t^k_{enc}$.

\paragraph{Text-Video Retrieval.} 
Contrastive learning alone has recently been found to learn better representations than its predictive counterpart in multi-modal pre-training~\citep{clip}. 
When used together with the predictive counterpart~\citep{li2021align}, it also boosts the performance. To align the video encoder and text encoder, we also utilize 
both the contrastive and predictive learning objectives. We apply contrastive learning over the output representations of the video encoder and the text encoder by optimizing a symmetric InfoNCE loss.
The video-to-text contrastive loss $\mathcal{L}_{v2t}$ is:
\begin{equation*}
    \mL_{v2t} = -\frac{1}{B} \sum^B_{k=1} \log \frac{\exp(\v^{k\top}_{cls}\t^{k}_{cls}/ \tau_c)}{\sum_j \exp(\v^{k\top}_{cls}\t^{j}_{cls}/ \tau_c)},
\end{equation*}
where  $\tau_c$ is a learnable temperature parameter initialized as 0.07.
Similarly, the text-to-video contrastive loss
$\mathcal{L}_{t2v}$ is:
\begin{equation*}
    \mL_{t2v} = -\frac{1}{B} \sum^B_{k=1} \log \frac{\exp(\v^{k\top}_{cls}\t^{k}_{cls}/ \tau_c)}{\sum_j \exp(\v^{j\top}_{cls}\t^{k}_{cls}/ \tau_c)}.
\end{equation*}
The video-text contrastive loss is defined as:
\begin{equation}
    \mL_{\mathrm{vtc}}=\frac{1}{2}(\mL_{v2t}+\mL_{t2v}).
\end{equation}
Following~\citet{li2021align}, besides the contrastive loss, we also use a video-text matching
loss  $\mL_{\mathrm{vtm}}$,
which predicts whether a pair of video and text is matched or not. 
For the $k$-th video-text pair, 
we map the joint video-text embedding $\t_{enc}^k$ to a two-class probability $\p_{vtm}^k$, and calculate $\mL_{\mathrm{vtm}}$ as:
\begin{equation}
    \mL_{\mathrm{vtm}} = \frac{1}{B} \sum^B_{k=1} \mathrm{CE}(\y_{vtm}^k, \p_{vtm}^k),
    \label{eq:matching_loss}
\end{equation}
where $\y_{vtm}^k$ is a 2-dimensional one-hot vector representing the ground-truth label, and $ \mathrm{CE}(\cdot,\cdot)$ is cross-entropy loss. 
The in-batch negatives used for $\mL_{\mathrm{vtm}}$  are mined based on the contrastive similarity
following~\citet{albef}.
The overall training objective is:
\begin{equation}
\mL_{\mathrm{retrieval}} = 
\mL_{\mathrm{vtc}}+
\mL_{\mathrm{vtm}}. \label{eq:retrieval}
\end{equation}

\paragraph{Video Question Answering. 
} 
Following \citet{DBLP:conf/cvpr/FanZZW0H19}, we view video question answering as a multimodal classification task and add a two-layer MLP classification head on the multimodal $\texttt{[Encode]}$ token. 
Assume the number of answers is $K$, 
for the $k$-th video-text pair,
we map the joint video-text embedding $\t_{enc}^k$ to a $K$-class probability $\p_{ans}^k$, the training objective for video question answering is:
\begin{equation}
    \mL_{\mathrm{VideoQA}} = \frac{1}{B} \sum^B_{k=1} \mathrm{CE}(\y_{ans}^k, \p_{ans}^k),
    \label{eq:vqa}
\end{equation}
where  $\y_{ans}^k$ is a $K$-dimensional one-hot classification label. 

\section{Experiments}
\label{sec:expt}
In this section, we evaluate the efficacy of the proposed LiteVL on the text-video retrieval and video question answering (VideoQA) tasks.
We initialize the weights of LiteVL from BLIP ~\citep{blip}, which uses a  ViT-B/16 as the image encoder, a BERT$_\mathrm{base}$ as the text encoder with additional cross-attention layers for the image-grounded text encoder.
We use both BLIP variants pre-trained on 14M and 129M image-text pairs, respectively, and the corresponding LiteVL initialized from them are dubbed as
LiteVL$_S$  and  LiteVL$_L$, respectively.

During fine-tuning, we randomly sample 8 and 16 frames per video for retrieval and VideoQA tasks, respectively. While in the inference stage, the frames are uniformly sampled. 
Following previous works~\citep{DBLP:journals/corr/abs-2201-04850,gorti2022xpool} and BLIP's pre-training setting, we resize each of the raw frames to 224$\times$224 before feeding them into the model. For the text-dependent pooling, the temperature $\tau$ in Eq.(\ref{eq:gamma}) is set to 1.0 by default.
More detailed training details and hyperparameters are in Appendix \ref{expt:appendix-setup}. 


\subsection{Text-Video Retrieval}

\paragraph{Datasets and Metrics.} 
We finetune on two text-video retrieval datasets: (i)
\textbf{MSRVTT}~\citep{DBLP:conf/cvpr/XuMYR16} 
consists of 10k videos and 200k text captions. 
Each video is paired with about 20 manually-labeled captions, and lasts about 10 to 32 seconds. 
There are two widely used ways to split the dataset, i.e., \textbf{MSRVTT}-7k \citep{howto100m} and \textbf{MSRVTT}-9k \citep{DBLP:conf/eccv/Gabeur0AS20}, 
which have 7k videos and 9k videos for training, respectively.
For a  comprehensive comparison with previous works, we use both splits that share the same 1k testing videos \citep{DBLP:conf/iccv/BainNVZ21}.  
(ii) \textbf{DiDeMo}~\citep{DBLP:conf/iccv/HendricksWSSDR17} consists of 10k Flickr videos annotated with 40k text captions. 
We evaluate text-video retrieval following \citet{DBLP:conf/cvpr/LeiLZGBB021}, 
where all captions for the same video are concatenated into a single query.

We evaluate text-video retrieval by R@k and MdR following~\citet{DBLP:conf/iccv/BainNVZ21}.
R@k means the recall (\%) through k retrieval efforts.
MdR represents the median rank of the retrieved video. 


\paragraph{Comparison with BLIP.}
Previous BLIP concatenates the image features of all frames as the aggregated video feature and feeds it to the image-grounded text encoder. 
In Table~\ref{tab:ablation_timesformer}, we show the comparison between our proposed LiteVL and the original BLIP as well as its variants with increased resolution (i.e., 384$\times$384), pre-training on COCO~\citep{lin2014microsoft} and fine-tuning setting.
As can be seen,
though inherited from the BLIP model,
our proposed  LiteVL clearly outperforms the original BLIP due to the explicit temporal modeling in both model-wise and feature-wise manners.
In particular, LiteVL$_S$ improves the R1 of the best-performed BLIP (14M) variant by 2.0, 2.6, and 2.4 points on \textbf{MSRVTT}-7k, \textbf{MSRVTT}-9k, and \textbf{DiDeMo}, respectively.

\begin{table}[ht!]
\centering
\scalebox{0.74}{
    \begin{tabular}{lcccc}
    \midrule [\heavyrulewidth]
    \textbf{Methods} &\textbf{R1}$\uparrow$& \textbf{R5}$\uparrow$& \textbf{R10}$\uparrow$& \textbf{MdR}$\downarrow$\\
    \midrule [\heavyrulewidth]
    NoiseEST~\citep{DBLP:conf/aaai/AmraniBRB21} & 17.4& 41.6& 53.6& 8 \\ 
    ClipBERT~\citep{DBLP:conf/cvpr/LeiLZGBB021} & 22.0& 46.8& 59.9& 6 \\
    VideoClip~\citep{DBLP:conf/emnlp/XuG0OAMZF21} & 30.9& 55.4& 66.8& - \\
    ALPRO~\citep{li2021align} & 33.9& 60.7& 73.2& 3\\
     CLIP4Clip~\citep{DBLP:journals/corr/abs-2104-08860} & 42.1& 71.9& 81.4 &2 \\
    BLIP$^\dagger$~\citep{blip} & 43.3 & 65.6& 74.7 & 2\\
    X-Pool~\citep{gorti2022xpool} & 43.9& 72.5& 82.3& 2\\
    \midrule
     LiteVL$_S$   &44.5 &70.3 &80.2&2 \\
     LiteVL$_L$   &\textbf{48.9} &\textbf{74.5} &\textbf{83.6} &\textbf{2} \\
    \midrule [\heavyrulewidth]
    \end{tabular}
    }
\caption{Results of
text-video retrieval on the test split of \textbf{MSRVTT}-7k. $\dagger$ means zero-shot results reported by the original BLIP paper.  
}
\label{tab:accents_msrvtt7k}
\end{table}

\begin{table}[ht!]
\centering
    \scalebox{0.75}{
    \begin{tabular}{lcccc}
    \midrule [\heavyrulewidth]
    \textbf{Methods} &\textbf{R1}$\uparrow$& \textbf{R5}$\uparrow$& \textbf{R10}$\uparrow$& \textbf{MdR}$\downarrow$\\
    \midrule [\heavyrulewidth]
    CE~\citep{DBLP:conf/bmvc/LiuANZ19} & 20.9& 48.8& 62.4& 6 \\ 
    Frozen~\citep{DBLP:conf/iccv/BainNVZ21} & 31.0&59.5& 70.5& 3 \\
    BridgeFormer~\citep{DBLP:journals/corr/abs-2201-04850} & 37.6&64.8& 75.1& 3 \\
    CLIP4Clip~\citep{DBLP:journals/corr/abs-2104-08860} & 44.5& 71.4& 81.6& 2 \\
    BLIP$^\dagger$~\citep{blip} & 43.3 & 65.6& 74.7 & 2\\
    X-Pool~\citep{gorti2022xpool} & 46.9& 72.8& 82.2& 2 \\
    \midrule
    LiteVL$_S$   &46.7&71.8 &81.7&2 \\
    LiteVL$_L$   &\textbf{50.8} &\textbf{76.3} &\textbf{84.4} &\textbf{1}\\
    \midrule [\heavyrulewidth]
    \end{tabular}
        }
\caption{Results of text-video retrieval on the test split of \textbf{MSRVTT}-9k dataset.}
\label{tab:accents_msrvtt9k}
\end{table}

\begin{table}[ht!]
\centering
\scalebox{0.75}{
    \begin{tabular}{lcccc}
    \midrule [\heavyrulewidth]
    \textbf{Methods} &\textbf{R1}$\uparrow$& \textbf{R5}$\uparrow$& \textbf{R10}$\uparrow$& \textbf{MdR}$\downarrow$\\
    \midrule [\heavyrulewidth]
    CE~\citep{DBLP:conf/bmvc/LiuANZ19} & 16.1& 41.1& 82.7 & 8 \\ 
    ClipBERT~\citep{DBLP:conf/cvpr/LeiLZGBB021} & 20.4& 48.0& 60.8 & 6 \\
    Frozen~\citep{DBLP:conf/iccv/BainNVZ21} & 34.6& 65.0& 74.7 & 3 \\
    ALPRO~\citep{li2021align} & 35.9& 67.5& 78.8 & 3 \\
    BridgeFormer~\citep{DBLP:journals/corr/abs-2201-04850} & 37.0& 62.2& 73.9 &3 \\
     CLIP4Clip~\citep{DBLP:journals/corr/abs-2104-08860} & 43.4& 70.2& 80.6 &2 \\
    \midrule
    LiteVL$_S$  &\textbf{53.7} &79.6 & 87.0 &1 \\
    LiteVL$_L$  &53.4&\textbf{80.7}&\textbf{87.0}&\textbf{1} \\
    \midrule [\heavyrulewidth]
    \end{tabular}
    }
\caption{Results of text-video retrieval on the test split of \textbf{DiDeMo} dataset.}
\label{tab:accents_didemo}
\end{table}

\paragraph{Comparison with Other Methods.}
Table~\ref{tab:accents_msrvtt7k}, Table~\ref{tab:accents_msrvtt9k} and Table \ref{tab:accents_didemo}
show the comparison between LiteVL and recent methods on text-video retrieval 
on \textbf{MSRVTT}-7k, \textbf{MSRVTT}-9k and \textbf{DiDeMo}, respectively. 
On all three datasets, LiteVL surpasses  previous works by a clear margin, including
methods requiring heavy video-text pre-training (e.g., ALPRO, CLIP4Clip and BridgeFormer) or 
based on image-language pre-trained models (e.g., ClipBERT and X-Pool).
Note that X-Pool also uses a parametric text-dependent pooling method to aggregate video features from an image-language pre-trained model. However, our proposed LiteVL still outperforms it with the simpler non-parametric similarity-based text-dependent pooling and the proposed dynamic scaling.
Besides, 
LiteVL$_L$ performs significantly  better than
LiteVL$_S$ on both splits of MSRVTT, and similarly on DiDeMo. 
This  indicates that the general multimodal priors learned from a large-scale image-text dataset can also benefit text-video retrieval tasks.


\subsection{Video Question Answering}

\paragraph{Datasets and Metrics.}
(i) \textbf{MSRVTT-QA}~\citep{DBLP:conf/mm/XuZX0Z0Z17}  consists of 10k videos, 243k open-ended questions, 
and 1.5k answer candidates based on MSRVTT.
(ii)  \textbf{MSVD-QA}~\citep{DBLP:conf/mm/XuZX0Z0Z17} is based on MSVD~\citep{DBLP:conf/acl/ChenD11}, and includes 1,970 videos, 50k question-answer pairs, and 2,423 answer candidates. 
We use top-1 accuracy (\%) as the evaluation metric.
For both datasets, we use the same train/val/test splits as \citet{DBLP:conf/mm/XuZX0Z0Z17}, and
select the model with the highest accuracy on the validation set for testing.

\begin{table}[htbp]
\centering
\scalebox{0.75}{
    \begin{tabular}{lcc}
    \midrule [\heavyrulewidth]
    \textbf{Methods}& \textbf{MSRVTT-QA}& \textbf{MSVD-QA}\\
    
    \midrule [\heavyrulewidth]
    HME~\citep{DBLP:conf/cvpr/FanZZW0H19} & 33.0& 33.7\\ 
    HGA~\citep{DBLP:conf/aaai/JiangH20}&35.5&34.7\\
    ClipBERT \citep{DBLP:conf/cvpr/LeiLZGBB021}&37.4&-\\
    SSML~\citep{DBLP:conf/aaai/AmraniBRB21}&35.1&35.1\\
    CoMVT~\citep{DBLP:conf/cvpr/SeoNS21}&39.5&42.6\\
    VQA-T~\citep{DBLP:conf/iccv/YangMSLS21} &41.5&46.3\\
    ALPRO~\citep{li2021align}&42.1&45.9\\
    \midrule
    LiteVL$_S$  (Ours)  &\textbf{42.9} &\textbf{47.3}  \\
   LiteVL$_L$  (Ours) & 42.5 &42.9\\ 
    \midrule [\heavyrulewidth]
    \end{tabular}
    }
\caption{Top-1 accuracy (\%) of VideoQA  on the test set of  \textbf{MSRVTT-QA}
and \textbf{MSVD-QA}.}
\label{tab:accents_qa}
\end{table}

\paragraph{Comparison with Other Methods.}
In Table~\ref{tab:accents_qa}, we compare our proposed LiteVL with existing methods on the open-ended video question answering datasets \textbf{MSRVTT-QA} and \textbf{MSVD-QA}. 
Again, LiteVL 
outperforms all the compared methods, including
VQA-T~\citep{DBLP:conf/iccv/YangMSLS21} pre-trained with 69M VideoQA samples
and the recent ALPRO pre-trained with 5.5M video-text pairs.
Instead, LiteVL  requires no general or QA-related video-text pre-training.  
Interestingly, LiteVL$_L$ is initialized from BLIP pre-trained on more image-text pairs than  LiteVL$_S$, but performs worse on two VideoQA tasks. 
We speculate this is because 
the additional 115M image-text pairs used to train BLIP (129M)  crawled from the web or filtered by BLIP captioner  are noisy and may bias towards the captioner. This may mislead 
the VideoQA task which requires more precise multimodal reasoning to answer questions than retrieval task.

\subsection{Ablation Study}
\begin{table*}[htbp]
\centering
\scalebox{0.74}{
	\begin{tabular}{l c c c c c c c c c}
		\toprule
        \multirow{2}{*}[-3pt]{\textbf{Methods}} &  \multicolumn{4}{c}{\textbf{MSRVTT}-7k} & \multicolumn{4}{c}{\textbf{DiDeMo}} & \multicolumn{1}{c}{\textbf{MSVD-QA}} \\ 
		\cmidrule(lr){2-5} \cmidrule(lr){6-9}  &\textbf{R1}$\uparrow$& \textbf{R5}$\uparrow$& \textbf{R10}$\uparrow$& \textbf{MdR}$\downarrow$ &\textbf{R1}$\uparrow$& \textbf{R5}$\uparrow$& \textbf{R10}$\uparrow$& \textbf{MdR}$\downarrow$& \textbf{Acc.}$\uparrow$  \\
		\midrule [\heavyrulewidth]
        LiteVL$_S$  &\textbf{44.5}&
        \textbf{70.3}&\textbf{80.2}&\textbf{2}&\textbf{53.7}&79.6&\textbf{87.0}&\textbf{1}&\textbf{47.3} \\
		\midrule
		     \multicolumn{10}{c}{\textit{Dynamic Temporal Scaling}} \\ \midrule
		\quad w/o Scaling ($\alpha_{t,l}=1$) & {44.4} & {69.4} & {79.5} & {2} & {52.8} & \textbf{80.3}& 87.0 & {1}& {47.1} \\
		\quad w/o Temporal attention ($\alpha_{t,l}=0$)  &42.8&68.6&79.4&2&51.7&78.5&85.6&1&45.5 \\
	    \midrule
     \multicolumn{10}{c}{\textit{Text-dependent Pooling}} \\ \midrule
		  \quad Original features  & 42.8 & 68.8& 79.8& 2 & 51.4 & 77.9& 85.7 & 1& 45.3 \\
		  \quad Original features + Temporal pooling   &43.0 &69.5 &79.8  &2  & 52.2 & 79.1&86.5  &1 &46.3  \\
		  \quad Original features + Spatial pooling   & 43.0 & 69.1 & 79.5 & 2 & 51.7 &78.9& 85.9 & 1& 45.9 \\
		  \quad Original features + Temporal pooling + Spatial pooling   & {43.6} & {69.8} & {79.8} & {2} & {52.3} & {79.3}& {86.5}& {1}& {46.4} \\

\midrule [\heavyrulewidth]
	\end{tabular}}

	\caption{Ablation studies on dynamic temporal scaling and text-dependent pooling of the proposed LiteVL. 
    } 
    \label{tab:ablation_scaling_pooling}
    
\end{table*}

\label{expt:ablation}

\subsubsection{Dynamic Temporal Scaling}
In Table~\ref{tab:ablation_scaling_pooling}, we compare the proposed dynamic temporal scaling against using
(i) constant scaling $\alpha_{l,t}=1$ in Eq.(\ref{equ:alt}), which reduces to directly using TimeSformer as the video encoder; and
(ii) constant scaling $\alpha_{l,t}=0$ in Eq.(\ref{equ:alt}), which reduces to using the ViT image encoder as the video encoder.
As can be seen, 
the proposed dynamic scaling learned upon each task performs better than the two special cases. 
By adopting TimeSformer ($\alpha_{l,t}=1$) instead of ViT ($\alpha_{l,t}=0$), the performance is boosted since the temporal dependency is  considered via the additional temporal attention module. 
With the proposed lightweight temporal scaling to adjust the 
frame-level importance according to each specific task,
the performance is further improved.

\begin{figure}[t!]
    \centering
    \begin{subfigure}{0.233\textwidth}
        \centering
  \includegraphics[width=\textwidth]{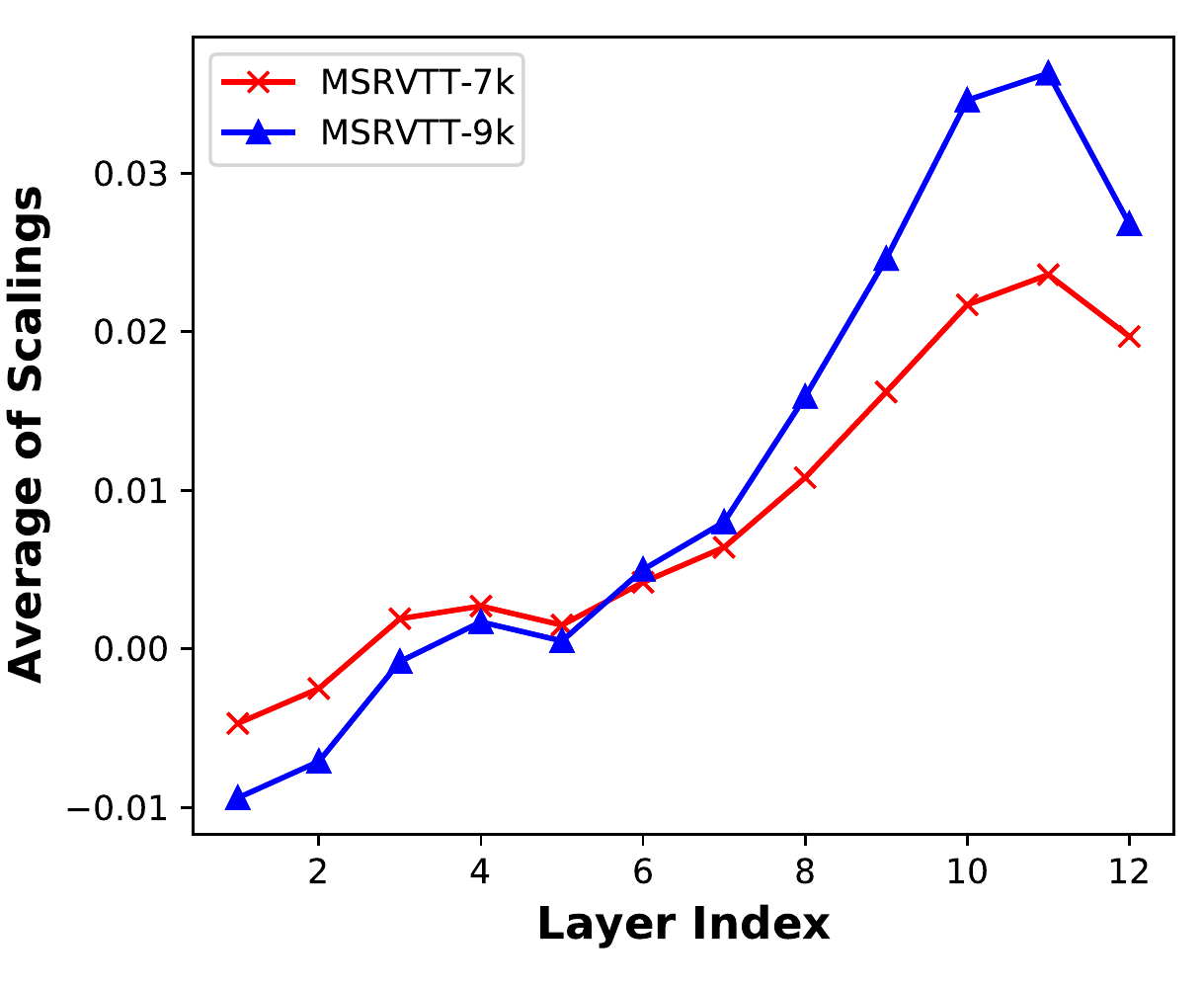}
        \caption{Text-video retrieval.}
        \label{fig:retrieval_scaling}
    \end{subfigure}
    \begin{subfigure}{0.237\textwidth}
        \centering
        \includegraphics[width=\textwidth]{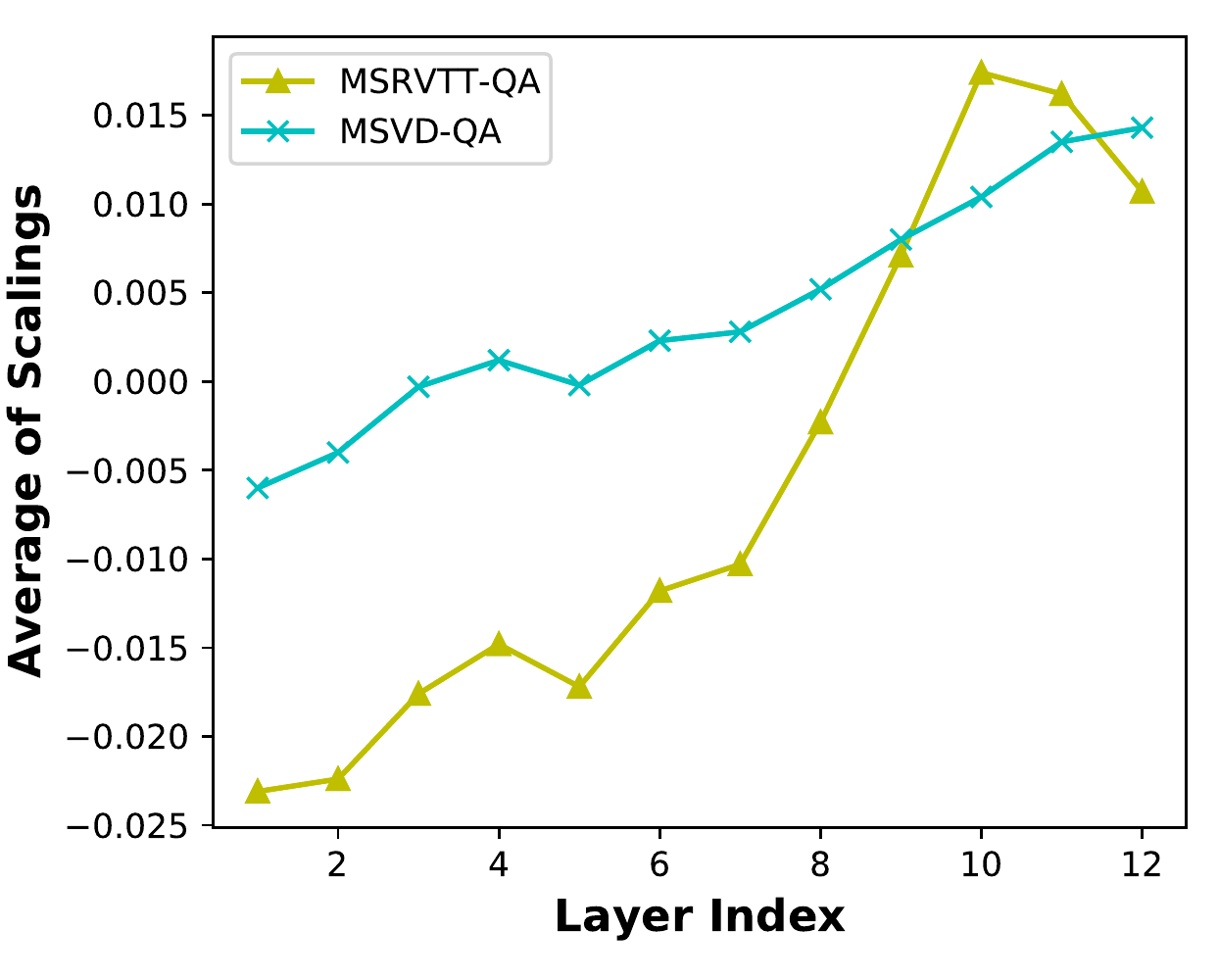}
        \caption{VideoQA.}
        \label{fig:videoqa_scaling}
    \end{subfigure}     
      \caption{Average temporal scalings $\gamma_{l,t}$  for different frames of each layer in the video encoder of LiteVL$_S$ trained on different tasks.
      }
    \label{fig:scalingfig}
\end{figure}

\paragraph{Visualization.}
As shown in  Figure~\ref{fig:scalingfig}, 
we visualize the average of the learned scalings $\gamma_{l,t}$ of each layer in the video encoder for both retrieval  (i.e., \textbf{MSRVTT}-7k, \textbf{MSRVTT}-9k) and VideoQA (i.e., \textbf{MSRVTT-QA}, \textbf{MSVD-QA}) tasks. 
For all datasets,
the average scaling is lower than 0 at the first layer and then shows an upward trend as the depth increases.
This indicates that
the shallow layers 
focus more on understanding the content of each frame, and pay less attention to temporal dependency among different frames. When the depth increases,
the spatial feature of each frame becomes more global
~\citep{vit}, and
the model gradually seeks to learn the temporal dependencies among them.

\subsubsection{Text-dependent Pooling}
\label{sec:ablation-pooling}
In Table~\ref{tab:ablation_scaling_pooling}, we compare our proposed text-dependent pooling in Section~\ref{sec:pooling} against several baseline pooling methods using different combinations of the original features $\V_f$, spatially pooled features $\V_{f_t}$ and temporally pooled features $\V_{f_s}$. 
As can be seen, compared with using only the original features, using either additional spatially or temporally pooled features improves the performance, 
and combining both of them further boosts performance.  
When coupled with the reweighting mechanism in Section~\ref{sec:pooling}, 
our proposed LiteVL obtains the best performance.

In addition, since the visually or temporally pooled features have much smaller sizes than the original features, using them merely increases the computation or memory cost of the cross-attention module of the video-grounded text encoder. The extra computation or memory cost incurred here is theoretically relatively acceptable.

\paragraph{Effect of  Temperature in the Text-dependent Pooling.}
We vary the temperature $\tau$ between 0.01 and 5.0 of the text-dependent pooling to study the effect of the temperature in the text-dependent pooling. 
As is shown in Table~\ref{tab:ablation_tau}, 
when $\tau$ equals 1.0,  both text-video retrieval and video question answering 
achieve the best performance. 
Therefore, the temperature $\tau$ of this pooling method is set to 1.0 for all datasets by default.

\begin{table*}[htb]
\small
\centering
	\begin{tabular}{l c c c c c c c c c c }
		\toprule

        \multirow{2}{*}[-3pt]{\textbf{Pooling Methods}} &  \multirow{2}{*}[-3pt]{$\tau$}
		&  \multicolumn{4}{c}{\textbf{MSRVTT}-7k} & \multicolumn{4}{c}{\textbf{DiDeMo}} & \multicolumn{1}{c}{\textbf{MSVD-QA}} \\ 
		\cmidrule(lr){3-6} \cmidrule(lr){7-10}  & &\textbf{R1}$\uparrow$& \textbf{R5}$\uparrow$& \textbf{R10}$\uparrow$& \textbf{MdR}$\downarrow$ &\textbf{R1}$\uparrow$& \textbf{R5}$\uparrow$& \textbf{R10}$\uparrow$& \textbf{MdR}$\downarrow$& \textbf{Acc.}$\uparrow$  \\
		\midrule [\heavyrulewidth]
		 &0.01& 8.2 & 24.7 & 35.1 & 23 & 6.7 & 23.5& 34.2 & 32& 27.4 \\
	    \multirow{3}{*}{Text-dependent pooling}&0.1& 44.3 & 69.7 & 79.0 & 2 & 52.1 & 80.0& 86.9 & 1& 46.1 \\
	    &1.0& \textbf{44.4} & \textbf{69.4} & \textbf{79.5} & \textbf{2} & \textbf{52.8} & \textbf{80.3}& \textbf{87.0} & \textbf{1}& \textbf{47.1} \\
	 &2.0& 43.9 &69.1 & 79.0 & 2 & 52.3 &79.8& 86.1& 1& 46.3 \\
	 &5.0& 44.1 & 69.3& 78.7 & 2 & 50.7& 79.6&86.3& 1& 46.4 \\
\midrule [\heavyrulewidth]

	\end{tabular}
	\caption{ Effect of different temperatures ($\tau$) in the text-dependent pooling on LiteVL$_S$. 
    }
    \label{tab:ablation_tau}
\end{table*}

\begin{table*}[htbp!]
\centering
\scalebox{0.85}{
	\begin{tabular}{l c c c c c c c c c ccc}
		\toprule
        \multirow{2}{*}[-3pt]{\textbf{Methods}} &  \multicolumn{4}{c}{\textbf{MSRVTT}-7k} &\multicolumn{4}{c}{\textbf{MSRVTT}-9k} & \multicolumn{4}{c}{\textbf{DiDeMo}} \\ 
		\cmidrule(lr){2-5} \cmidrule(lr){6-9} \cmidrule(lr){10-13} &\textbf{R1}$\uparrow$& \textbf{R5}$\uparrow$& \textbf{R10}$\uparrow$& \textbf{MdR}$\downarrow$
		&\textbf{R1}$\uparrow$& \textbf{R5}$\uparrow$& \textbf{R10}$\uparrow$& \textbf{MdR}$\downarrow$
		&\textbf{R1}$\uparrow$& \textbf{R5}$\uparrow$& \textbf{R10}$\uparrow$& \textbf{MdR}$\downarrow$ \\
		\midrule [\heavyrulewidth]
        LiteVL$_S$  &\textbf{44.5}&
        \textbf{70.3}&\textbf{80.2}&\textbf{2}&\textbf{46.7}&71.8&\textbf{81.7}&\textbf{2}&\textbf{53.7}&\textbf{79.6}&87.0&\textbf{1}\\
        \quad w/ top-$k$ &44.3&69.9&80.1&2&46.4&\textbf{72.0}&81.1&2&53.5&79.5&\textbf{87.1}&1\\
\midrule    
   LiteVL$_L$  &\textbf{48.9} &74.5
         &\textbf{83.6} &\textbf{2} & \textbf{50.8}&76.3&\textbf{84.4} &\textbf{1} &\textbf{53.4} & \textbf{80.7}& \textbf{87.0}& \textbf{1}\\     
        \quad w/ top-$k$ &48.4&\textbf{75.4}&82.5&2& 50.5&\textbf{76.5}& 84.1&1 &52.3&79.2&86.8&1\\
\midrule [\heavyrulewidth]
	\end{tabular}}
	\caption{Effect of using $s_\mathrm{vtc}$ to filter top-$k$ ($k$=100) candidates and calculate their $s_\mathrm{vtm}$ score for ranking. 
    } 
    \label{tab:top-$k$}
    
\end{table*}

\begin{figure}[ht!]
    \centering
    \includegraphics[width=1\linewidth]{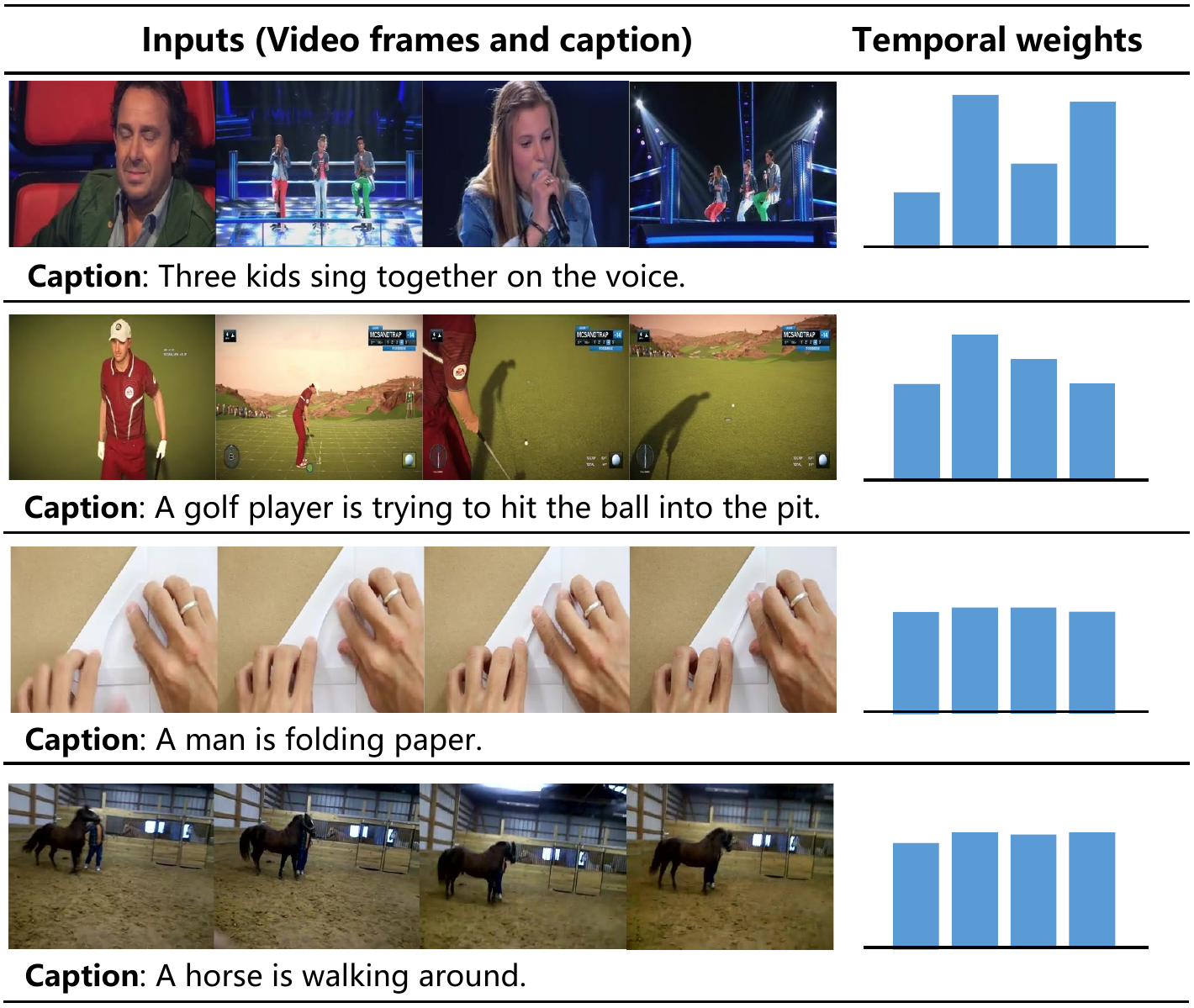}
    \caption{Bar plots of temporal weights in text-dependent pooling. 
    Video frames more related to the caption are
    assigned with higher weights.}
	\label{fig:visual}
\end{figure}

\paragraph{Visualization.}
To better understand the effect of text-dependent pooling, 
we use LiteVL$_S$ to visualize the video-text pair from \textbf{MSRVTT}-7k testing set and their corresponding temporal weights
 (${\bf g}_t$ in Eq.(\ref{eq:gamma})).
As show in Figure~\ref{fig:visual}, when the changes among different frames are relatively large,
the proposed text-dependent pooling encourages the model to assign higher weights to  frames better described by the caption. 
For instance, in the first example, the second and fourth frames are more related to the caption \texttt{``Three kids sing together on the voice.''} and assigned higher weights.

\begin{figure}[ht!]
    \centering
    \includegraphics[width=1\linewidth]{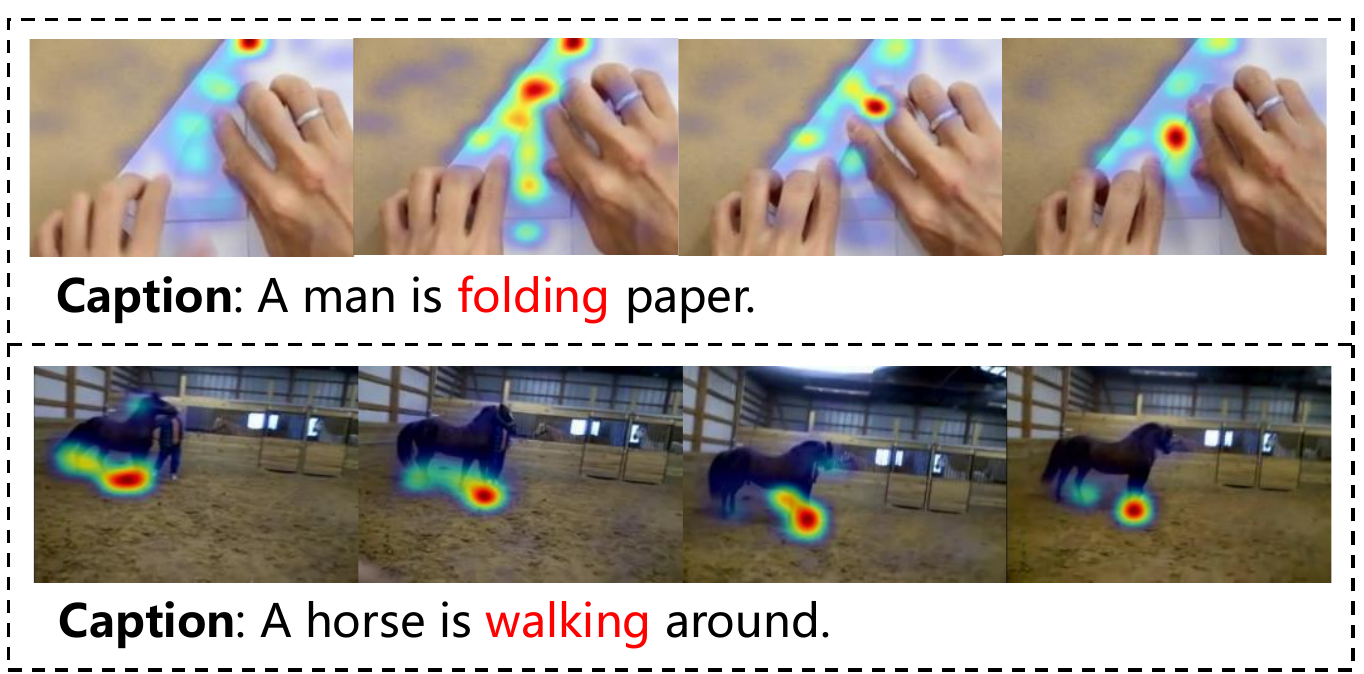}
    \caption{Grad-CAM visualizations on the cross-attention maps corresponding to highlighted keywords.
    }
	\label{fig:visual_gradcam}
\end{figure}

On the contrary, as can be seen from the last two examples in Figure~\ref{fig:visual}, when the different frames  only differ in minor changes and  each frame is similarly close to the caption, the learned weights for each frame are also similar. 
For these cases, we further study the more fine-grained spatial-temporal dependencies using the  Grad-CAM~\citep{selvaraju2017grad} visualizations.
We compute Grad-CAM using the cross-attention maps averaged 
over all attention heads in the 8-th layer (a specialized layer in grounding) of the video-grounded text encoder.
The gradients are acquired by maximizing the video-text matching score in Eq.(\ref{eq:matching_loss}).
As can be seen in Figure~\ref{fig:visual_gradcam}, 
the proposed
LiteVL effectively captures the minor changes among different frames.
This also indicates that our proposed
text-dependent pooling 
provides fruitful information for the video-grounded text encoder.
More visualizations are
in Appendix~\ref{appendix_visualization}.


\section{Discussions}
\subsection{Extension to Other Image-language 
Pre-trained Models}

In this work, we choose BLIP to initialize our proposed model mainly because (i) it performs well on various downstream image-language tasks; and (ii) it can be regarded as a single-stream and dual-stream hybrid structure.
Its dual-stream part allows efficient inference for cross-modal retrieval tasks, while its cross-attention allows deep cross-modal interaction for tasks like VQA.

On the other hand, the proposed dynamic temporal scaling and text-dependent pooling can also be applied to the dual-stream model like CLIP~\citep{clip}. 
For this setting, we also conduct a simple experiment.
For CLIP, we use the proposed text-dependent pooling on top of the video features. 
As CLIP relies on the global features for retrieval, instead of concatenation in Eq.(\ref{eq:concat_video_feature}), we compute a weighted average of the reweighted features. 
Compared with a recent work CLIP4Clip which also extends CLIP for video retrieval,
CLIP with our proposed method improves the best CLIP4Clip-meanP method by 1.9\% and 1.7\%, for the R1 and R10 on \textbf{MSRVTT}-7k, respectively. 

\subsection{Scaling to Larger-scale Retrieval Tasks}
\label{sec:topkfilter}

Since the test set sizes for all three retrieval datasets used in this work are relatively small, we compute a pairwise VTM score $s_\mathrm{vtm}$ for all video-text pairs during inference.
However, the speed of inference in this approach will be slow when the size of dataset is huge in real-world scenarios.

In this section, we provide a more efficient retrieval solution.
Specifically, we first compute the video-text similarity score $s_\mathrm{vtc}$ for all video-text pairs, then we take the top-$k$ candidates and calculate their VTM score for ranking. Such method can speed up inference, because the $k$ can be set to be very small compared with the test set size.
 Table~\ref{tab:top-$k$} shows that using this efficient two-stage retrieval solution has negligible performance degradation.
 

\section{Conclusion}
\label{sec:bibtex}
We propose LiteVL, 
a video-language model without  heavy video-language pre-training or object detectors. 
LiteVL inherits the already-learned alignment between the spatial visual information and textual information, 
from a pre-trained image-language  model. 
Then, an extra temporal attention with dynamic temporal scaling is proposed to learn the temporal dynamics in the video. 
We also introduce a non-parametric pooling method which aggregates video features conditioned on the text description, 
enabling fine-grained video-language alignment. 
Empirical results show that our LiteVL outperforms the state-of-the-art methods 
trained with much more training data.

\section*{Acknowledgements}
We gratefully acknowledge the support of ~\citet{mindspore2021}\footnote{\url{https://www.mindspore.cn/}} and Ascend AI Processor for this research.
\section*{Limitations}
Although LiteVL achieves great performance without the video-language pre-training, 
it is interesting to explore how to combine a few
video-language pairs from the pre-training datasets
to  further boost the performance. 
In addition, we view VideoQA as a classification task, which may be limited to a fixed number of answer candidates.
In the future, we would like to generation ability 
of the proposed LiteVL 
on more diverse 
long-form visual question answering datasets.


\section*{Ethics Statement}
During the training process, the knowledge  from the inherited image-text pre-trained models and the training data may have unsuitable information. Therefore, our trained models are possibly exposed to the similar risks of large language models as mentioned in \citet{weidinger2021ethical}. Thanks to \citet{thoppilan2022lamda}, harmful training data can be filtered to improve the safety of our model. On the other hand, there is no generative task involved in our video-text understanding framework, which means our framework will not output offensive language.

Before deploying the models in real-world applications, risk analysis is necessary since the used data may contain some unintended privacy information and improper video/text that do not conform to social norms.

\bibliography{anthology,custom}

\begin{thebibliography}{43}
\expandafter\ifx\csname natexlab\endcsname\relax\def\natexlab#1{#1}\fi

\bibitem[{Amrani et~al.(2021)Amrani, Ben{-}Ari, Rotman, and
  Bronstein}]{DBLP:conf/aaai/AmraniBRB21}
Elad Amrani, Rami Ben{-}Ari, Daniel Rotman, and Alex~M. Bronstein. 2021.
\newblock \href {https://ojs.aaai.org/index.php/AAAI/article/view/16822} {Noise
  estimation using density estimation for self-supervised multimodal learning}.
\newblock In \emph{AAAI}.

\bibitem[{Arnab et~al.(2021)Arnab, Dehghani, Heigold, Sun, Lucic, and
  Schmid}]{vivit}
Anurag Arnab, Mostafa Dehghani, Georg Heigold, Chen Sun, Mario Lucic, and
  Cordelia Schmid. 2021.
\newblock \href {https://doi.org/10.1109/ICCV48922.2021.00676} {Vivit: {A}
  video vision transformer}.
\newblock In \emph{ICCV}.

\bibitem[{Bain et~al.(2021)Bain, Nagrani, Varol, and
  Zisserman}]{DBLP:conf/iccv/BainNVZ21}
Max Bain, Arsha Nagrani, G{\"{u}}l Varol, and Andrew Zisserman. 2021.
\newblock \href {https://doi.org/10.1109/ICCV48922.2021.00175} {Frozen in time:
  {A} joint video and image encoder for end-to-end retrieval}.
\newblock In \emph{ICCV}.

\bibitem[{Bertasius et~al.(2021)Bertasius, Wang, and Torresani}]{timesformer}
Gedas Bertasius, Heng Wang, and Lorenzo Torresani. 2021.
\newblock \href {http://proceedings.mlr.press/v139/bertasius21a.html} {Is
  space-time attention all you need for video understanding?}
\newblock In \emph{ICML}.

\bibitem[{Chen and Dolan(2011)}]{DBLP:conf/acl/ChenD11}
David~L. Chen and William~B. Dolan. 2011.
\newblock \href {https://aclanthology.org/P11-1020/} {Collecting highly
  parallel data for paraphrase evaluation}.
\newblock In \emph{ACL}.

\bibitem[{Chen et~al.(2020)Chen, Li, Yu, El~Kholy, Ahmed, Gan, Cheng, and
  Liu}]{chen2020uniter}
Yen-Chun Chen, Linjie Li, Licheng Yu, Ahmed El~Kholy, Faisal Ahmed, Zhe Gan,
  Yu~Cheng, and Jingjing Liu. 2020.
\newblock \href
  {https://www.ecva.net/papers/eccv_2020/papers_ECCV/papers/123750103.pdf}
  {Uniter: Universal image-text representation learning}.
\newblock In \emph{ECCV}.

\bibitem[{Cubuk et~al.(2020)Cubuk, Zoph, Shlens, and
  Le}]{DBLP:conf/nips/CubukZS020}
Ekin~Dogus Cubuk, Barret Zoph, Jonathon Shlens, and Quoc Le. 2020.
\newblock \href
  {https://proceedings.neurips.cc/paper/2020/hash/d85b63ef0ccb114d0a3bb7b7d808028f-Abstract.html}
  {Randaugment: Practical automated data augmentation with a reduced search
  space}.
\newblock In \emph{NeurIPS}.

\bibitem[{Devlin et~al.(2019)Devlin, Chang, Lee, and
  Toutanova}]{DBLP:conf/naacl/DevlinCLT19}
Jacob Devlin, Ming{-}Wei Chang, Kenton Lee, and Kristina Toutanova. 2019.
\newblock \href {https://doi.org/10.18653/v1/n19-1423} {{BERT:} pre-training of
  deep bidirectional transformers for language understanding}.
\newblock In \emph{NAACL}.

\bibitem[{Dosovitskiy et~al.(2021)Dosovitskiy, Beyer, Kolesnikov, Weissenborn,
  Zhai, Unterthiner, Dehghani, Minderer, Heigold, Gelly, Uszkoreit, and
  Houlsby}]{vit}
Alexey Dosovitskiy, Lucas Beyer, Alexander Kolesnikov, Dirk Weissenborn,
  Xiaohua Zhai, Thomas Unterthiner, Mostafa Dehghani, Matthias Minderer, Georg
  Heigold, Sylvain Gelly, Jakob Uszkoreit, and Neil Houlsby. 2021.
\newblock \href {https://openreview.net/forum?id=YicbFdNTTy} {An image is worth
  16x16 words: Transformers for image recognition at scale}.
\newblock In \emph{ICLR}.

\bibitem[{Fan et~al.(2019)Fan, Zhang, Zhang, Wang, Zhang, and
  Huang}]{DBLP:conf/cvpr/FanZZW0H19}
Chenyou Fan, Xiaofan Zhang, Shu Zhang, Wensheng Wang, Chi Zhang, and Heng
  Huang. 2019.
\newblock \href
  {http://openaccess.thecvf.com/content\_CVPR\_2019/html/Fan\_Heterogeneous\_Memory\_Enhanced\_Multimodal\_Attention\_Model\_for\_Video\_Question\_Answering\_CVPR\_2019\_paper.html}
  {Heterogeneous memory enhanced multimodal attention model for video question
  answering}.
\newblock In \emph{CVPR}.

\bibitem[{Gabeur et~al.(2020)Gabeur, Sun, Alahari, and
  Schmid}]{DBLP:conf/eccv/Gabeur0AS20}
Valentin Gabeur, Chen Sun, Karteek Alahari, and Cordelia Schmid. 2020.
\newblock \href {https://doi.org/10.1007/978-3-030-58548-8\_13} {Multi-modal
  transformer for video retrieval}.
\newblock In \emph{ECCV}.

\bibitem[{Ge et~al.(2022)Ge, Ge, Liu, Li, Shan, Qie, and
  Luo}]{DBLP:journals/corr/abs-2201-04850}
Yuying Ge, Yixiao Ge, Xihui Liu, Dian Li, Ying Shan, Xiaohu Qie, and Ping Luo.
  2022.
\newblock \href {https://doi.org/10.48550/arXiv.2203.15086} {Bridgeformer:
  Bridging video-text retrieval with multiple choice questions}.
\newblock In \emph{CVPR}.

\bibitem[{Gorti et~al.(2022)Gorti, Vouitsis, Ma, Golestan, Volkovs, Garg, and
  Yu}]{gorti2022xpool}
Satya~Krishna Gorti, Noel Vouitsis, Junwei Ma, Keyvan Golestan, Maksims
  Volkovs, Animesh Garg, and Guangwei Yu. 2022.
\newblock \href {https://doi.org/10.48550/arXiv.2203.15086} {X-pool:
  Cross-modal language-video attention for text-video retrieval}.
\newblock In \emph{CVPR}.

\bibitem[{Han et~al.(2021)Han, Xiao, Wu, Guo, Xu, and
  Wang}]{DBLP:conf/nips/HanXWGXW21}
Kai Han, An~Xiao, Enhua Wu, Jianyuan Guo, Chunjing Xu, and Yunhe Wang. 2021.
\newblock \href
  {https://proceedings.neurips.cc/paper/2021/hash/854d9fca60b4bd07f9bb215d59ef5561-Abstract.html}
  {Transformer in transformer}.
\newblock In \emph{NeurIPS}.

\bibitem[{Hendricks et~al.(2017)Hendricks, Wang, Shechtman, Sivic, Darrell, and
  Russell}]{DBLP:conf/iccv/HendricksWSSDR17}
Lisa~Anne Hendricks, Oliver Wang, Eli Shechtman, Josef Sivic, Trevor Darrell,
  and Bryan~C. Russell. 2017.
\newblock \href {https://doi.org/10.1109/ICCV.2017.618} {Localizing moments in
  video with natural language}.
\newblock In \emph{ICCV}.

\bibitem[{Jiang and Han(2020)}]{DBLP:conf/aaai/JiangH20}
Pin Jiang and Yahong Han. 2020.
\newblock \href {https://ojs.aaai.org/index.php/AAAI/article/view/6767}
  {Reasoning with heterogeneous graph alignment for video question answering}.
\newblock In \emph{AAAI}.

\bibitem[{Krishna et~al.(2016)Krishna, Zhu, Groth, Johnson, Hata, Kravitz,
  Chen, Kalantidis, Li, Shamma, Bernstein, and Fei{-}Fei}]{krishnavisualgenome}
Ranjay Krishna, Yuke Zhu, Oliver Groth, Justin Johnson, Kenji Hata, Joshua
  Kravitz, Stephanie Chen, Yannis Kalantidis, Li{-}Jia Li, David~A. Shamma,
  Michael~S. Bernstein, and Li~Fei{-}Fei. 2016.
\newblock \href {http://arxiv.org/abs/1602.07332} {Visual genome: Connecting
  language and vision using crowdsourced dense image annotations}.
\newblock \emph{CoRR}, abs/1602.07332.

\bibitem[{Lei et~al.(2021)Lei, Li, Zhou, Gan, Berg, Bansal, and
  Liu}]{DBLP:conf/cvpr/LeiLZGBB021}
Jie Lei, Linjie Li, Luowei Zhou, Zhe Gan, Tamara~L. Berg, Mohit Bansal, and
  Jingjing Liu. 2021.
\newblock \href
  {https://openaccess.thecvf.com/content/CVPR2021/html/Lei\_Less\_Is\_More\_ClipBERT\_for\_Video-and-Language\_Learning\_via\_Sparse\_Sampling\_CVPR\_2021\_paper.html}
  {Less is more: Clipbert for video-and-language learning via sparse sampling}.
\newblock In \emph{CVPR}.

\bibitem[{Li et~al.(2022{\natexlab{a}})Li, Li, Li, Niebles, and
  Hoi}]{li2021align}
Dongxu Li, Junnan Li, Hongdong Li, Juan~Carlos Niebles, and Steven C.~H. Hoi.
  2022{\natexlab{a}}.
\newblock \href {https://arxiv.org/abs/2112.09583} {Align and prompt:
  Video-and-language pre-training with entity prompts}.
\newblock In \emph{CVPR}.

\bibitem[{Li et~al.(2022{\natexlab{b}})Li, Li, Xiong, and Hoi}]{blip}
Junnan Li, Dongxu Li, Caiming Xiong, and Steven Hoi. 2022{\natexlab{b}}.
\newblock \href {https://arxiv.org/abs/2201.12086} {{BLIP:} bootstrapping
  language-image pre-training for unified vision-language understanding and
  generation}.
\newblock In \emph{ICML}.

\bibitem[{Li et~al.(2021)Li, Selvaraju, Gotmare, Joty, Xiong, and Hoi}]{albef}
Junnan Li, Ramprasaath~R. Selvaraju, Akhilesh Gotmare, Shafiq~R. Joty, Caiming
  Xiong, and Steven~Chu{-}Hong Hoi. 2021.
\newblock \href
  {https://proceedings.neurips.cc/paper/2021/hash/505259756244493872b7709a8a01b536-Abstract.html}
  {Align before fuse: Vision and language representation learning with momentum
  distillation}.
\newblock In \emph{NeurIPS}.

\bibitem[{Lin et~al.(2014)Lin, Maire, Belongie, Hays, Perona, Ramanan,
  Doll{\'{a}}r, and Zitnick}]{lin2014microsoft}
Tsung{-}Yi Lin, Michael Maire, Serge~J. Belongie, James Hays, Pietro Perona,
  Deva Ramanan, Piotr Doll{\'{a}}r, and C.~Lawrence Zitnick. 2014.
\newblock \href {https://doi.org/10.1007/978-3-319-10602-1\_48} {Microsoft
  {COCO:} common objects in context}.
\newblock In \emph{ECCV}.

\bibitem[{Liu et~al.(2019)Liu, Albanie, Nagrani, and
  Zisserman}]{DBLP:conf/bmvc/LiuANZ19}
Yang Liu, Samuel Albanie, Arsha Nagrani, and Andrew Zisserman. 2019.
\newblock \href {https://bmvc2019.org/wp-content/uploads/papers/0363-paper.pdf}
  {Use what you have: Video retrieval using representations from collaborative
  experts}.
\newblock In \emph{BMVC}.

\bibitem[{Liu et~al.(2021)Liu, Lin, Cao, Hu, Wei, Zhang, Lin, and
  Guo}]{DBLP:conf/iccv/LiuL00W0LG21}
Ze~Liu, Yutong Lin, Yue Cao, Han Hu, Yixuan Wei, Zheng Zhang, Stephen Lin, and
  Baining Guo. 2021.
\newblock \href {https://doi.org/10.1109/ICCV48922.2021.00986} {Swin
  transformer: Hierarchical vision transformer using shifted windows}.
\newblock In \emph{ICCV}.

\bibitem[{Luo et~al.(2020)Luo, Ji, Shi, Huang, Duan, Li, Chen, and
  Zhou}]{luo2020univl}
Huaishao Luo, Lei Ji, Botian Shi, Haoyang Huang, Nan Duan, Tianrui Li, Xilin
  Chen, and Ming Zhou. 2020.
\newblock \href {https://arxiv.org/abs/2002.06353} {Univilm: {A} unified video
  and language pre-training model for multimodal understanding and generation}.
\newblock \emph{CoRR}, abs/2002.06353.

\bibitem[{Luo et~al.(2021)Luo, Ji, Zhong, Chen, Lei, Duan, and
  Li}]{DBLP:journals/corr/abs-2104-08860}
Huaishao Luo, Lei Ji, Ming Zhong, Yang Chen, Wen Lei, Nan Duan, and Tianrui Li.
  2021.
\newblock \href {https://arxiv.org/abs/2104.08860} {Clip4clip: An empirical
  study of {CLIP} for end to end video clip retrieval}.
\newblock \emph{CoRR}, abs/2104.08860.

\bibitem[{Miech et~al.(2019)Miech, Zhukov, Alayrac, Tapaswi, Laptev, and
  Sivic}]{howto100m}
Antoine Miech, Dimitri Zhukov, Jean{-}Baptiste Alayrac, Makarand Tapaswi, Ivan
  Laptev, and Josef Sivic. 2019.
\newblock \href {https://doi.org/10.1109/ICCV.2019.00272} {Howto100m: Learning
  a text-video embedding by watching hundred million narrated video clips}.
\newblock In \emph{ICCV}.

\bibitem[{MindSpore()}]{mindspore2021}
MindSpore.
\newblock \url{https://www.mindspore.cn}.
\newblock 2021.

\bibitem[{Radford et~al.(2021)Radford, Kim, Hallacy, Ramesh, Goh, Agarwal,
  Sastry, Askell, Mishkin, Clark, Krueger, and Sutskever}]{clip}
Alec Radford, Jong~Wook Kim, Chris Hallacy, Aditya Ramesh, Gabriel Goh,
  Sandhini Agarwal, Girish Sastry, Amanda Askell, Pamela Mishkin, Jack Clark,
  Gretchen Krueger, and Ilya Sutskever. 2021.
\newblock \href {http://proceedings.mlr.press/v139/radford21a/radford21a.pdf}
  {Learning transferable visual models from natural language supervision}.
\newblock In \emph{ICML}.

\bibitem[{Selvaraju et~al.(2017)Selvaraju, Cogswell, Das, Vedantam, Parikh, and
  Batra}]{selvaraju2017grad}
Ramprasaath~R. Selvaraju, Michael Cogswell, Abhishek Das, Ramakrishna Vedantam,
  Devi Parikh, and Dhruv Batra. 2017.
\newblock \href {https://doi.org/10.1109/ICCV.2017.74} {Grad-cam: Visual
  explanations from deep networks via gradient-based localization}.
\newblock In \emph{ICCV}.

\bibitem[{Seo et~al.(2021)Seo, Nagrani, and Schmid}]{DBLP:conf/cvpr/SeoNS21}
Paul~Hongsuck Seo, Arsha Nagrani, and Cordelia Schmid. 2021.
\newblock \href
  {https://openaccess.thecvf.com/content/CVPR2021/html/Seo\_Look\_Before\_You\_Speak\_Visually\_Contextualized\_Utterances\_CVPR\_2021\_paper.html}
  {Look before you speak: Visually contextualized utterances}.
\newblock In \emph{CVPR}.

\bibitem[{Sharma et~al.(2018)Sharma, Ding, Goodman, and
  Soricut}]{sharma2018conceptual}
Piyush Sharma, Nan Ding, Sebastian Goodman, and Radu Soricut. 2018.
\newblock \href {https://aclanthology.org/P18-1238/} {Conceptual captions: A
  cleaned, hypernymed, image alt-text dataset for automatic image captioning}.
\newblock In \emph{ACL}.

\bibitem[{Sun et~al.(2019)Sun, Myers, Vondrick, Murphy, and
  Schmid}]{DBLP:conf/iccv/SunMV0S19}
Chen Sun, Austin Myers, Carl Vondrick, Kevin Murphy, and Cordelia Schmid. 2019.
\newblock \href {https://doi.org/10.1109/ICCV.2019.00756} {Videobert: {A} joint
  model for video and language representation learning}.
\newblock In \emph{ICCV}.

\bibitem[{Thoppilan et~al.(2022)Thoppilan, De~Freitas, Hall, Shazeer,
  Kulshreshtha, Cheng, Jin, Bos, Baker, Du et~al.}]{thoppilan2022lamda}
Romal Thoppilan, Daniel De~Freitas, Jamie Hall, Noam Shazeer, Apoorv
  Kulshreshtha, Heng-Tze Cheng, Alicia Jin, Taylor Bos, Leslie Baker, Yu~Du,
  et~al. 2022.
\newblock \href {https://arxiv.org/abs/2201.08239} {Lamda: Language models for
  dialog applications}.
\newblock \emph{CoRR}, abs/2201.08239.

\bibitem[{Touvron et~al.(2021)Touvron, Cord, Douze, Massa, Sablayrolles, and
  J{\'{e}}gou}]{DBLP:conf/icml/TouvronCDMSJ21}
Hugo Touvron, Matthieu Cord, Matthijs Douze, Francisco Massa, Alexandre
  Sablayrolles, and Herv{\'{e}} J{\'{e}}gou. 2021.
\newblock \href {http://proceedings.mlr.press/v139/touvron21a.html} {Training
  data-efficient image transformers {\&} distillation through attention}.
\newblock In \emph{ICML}.

\bibitem[{Vaswani et~al.(2017)Vaswani, Shazeer, Parmar, Uszkoreit, Jones,
  Gomez, Kaiser, and Polosukhin}]{DBLP:conf/nips/VaswaniSPUJGKP17}
Ashish Vaswani, Noam Shazeer, Niki Parmar, Jakob Uszkoreit, Llion Jones,
  Aidan~N. Gomez, Lukasz Kaiser, and Illia Polosukhin. 2017.
\newblock \href
  {https://proceedings.neurips.cc/paper/2017/hash/3f5ee243547dee91fbd053c1c4a845aa-Abstract.html}
  {Attention is all you need}.
\newblock In \emph{NeurIPS}.

\bibitem[{Wang et~al.(2021)Wang, Xie, Li, Fan, Song, Liang, Lu, Luo, and
  Shao}]{DBLP:conf/iccv/WangX0FSLL0021}
Wenhai Wang, Enze Xie, Xiang Li, Deng{-}Ping Fan, Kaitao Song, Ding Liang, Tong
  Lu, Ping Luo, and Ling Shao. 2021.
\newblock \href {https://doi.org/10.1109/ICCV48922.2021.00061} {Pyramid vision
  transformer: {A} versatile backbone for dense prediction without
  convolutions}.
\newblock In \emph{ICCV}.

\bibitem[{Weidinger et~al.(2021)Weidinger, Mellor, Rauh, Griffin, Uesato,
  Huang, Cheng, Glaese, Balle, Kasirzadeh et~al.}]{weidinger2021ethical}
Laura Weidinger, John Mellor, Maribeth Rauh, Conor Griffin, Jonathan Uesato,
  Po-Sen Huang, Myra Cheng, Mia Glaese, Borja Balle, Atoosa Kasirzadeh, et~al.
  2021.
\newblock \href {https://arxiv.org/abs/2112.04359} {Ethical and social risks of
  harm from language models}.
\newblock \emph{CoRR}, abs/2112.04359.

\bibitem[{Xu et~al.(2017)Xu, Zhao, Xiao, Wu, Zhang, He, and
  Zhuang}]{DBLP:conf/mm/XuZX0Z0Z17}
Dejing Xu, Zhou Zhao, Jun Xiao, Fei Wu, Hanwang Zhang, Xiangnan He, and Yueting
  Zhuang. 2017.
\newblock \href {https://doi.org/10.1145/3123266.3123427} {Video question
  answering via gradually refined attention over appearance and motion}.
\newblock In \emph{ACM MM}.

\bibitem[{Xu et~al.(2021)Xu, Ghosh, Huang, Okhonko, Aghajanyan, Metze,
  Zettlemoyer, and Feichtenhofer}]{DBLP:conf/emnlp/XuG0OAMZF21}
Hu~Xu, Gargi Ghosh, Po{-}Yao Huang, Dmytro Okhonko, Armen Aghajanyan, Florian
  Metze, Luke Zettlemoyer, and Christoph Feichtenhofer. 2021.
\newblock \href {https://doi.org/10.18653/v1/2021.emnlp-main.544} {Videoclip:
  Contrastive pre-training for zero-shot video-text understanding}.
\newblock In \emph{EMNLP}.

\bibitem[{Xu et~al.(2016)Xu, Mei, Yao, and Rui}]{DBLP:conf/cvpr/XuMYR16}
Jun Xu, Tao Mei, Ting Yao, and Yong Rui. 2016.
\newblock \href {https://doi.org/10.1109/CVPR.2016.571} {{MSR-VTT:} {A} large
  video description dataset for bridging video and language}.
\newblock In \emph{CVPR}.

\bibitem[{Yang et~al.(2021)Yang, Miech, Sivic, Laptev, and
  Schmid}]{DBLP:conf/iccv/YangMSLS21}
Antoine Yang, Antoine Miech, Josef Sivic, Ivan Laptev, and Cordelia Schmid.
  2021.
\newblock \href {https://doi.org/10.1109/ICCV48922.2021.00171} {Just ask:
  Learning to answer questions from millions of narrated videos}.
\newblock In \emph{ICCV}.

\bibitem[{Zhu and Yang(2020)}]{zhu2020actbert}
Linchao Zhu and Yi~Yang. 2020.
\newblock \href
  {https://openaccess.thecvf.com/content_CVPR_2020/papers/Zhu_ActBERT_Learning_Global-Local_Video-Text_Representations_CVPR_2020_paper.pdf}
  {Actbert: Learning global-local video-text representations}.
\newblock In \emph{CVPR}.

\end{thebibliography}
\bibliographystyle{acl_natbib}

\appendix

\newpage

\section{Detailed Fine-tuning Setups}
\label{expt:appendix-setup}

We list the detailed fine-tuning setups on each dataset in Table~\ref{tab:fine-tune_retrieval} and Table~\ref{tab:fine-tune_qa}. 
For all downstream datasets, we resize each frame to 224$\times$224 unless otherwise stated.
Following ALPRO~\citep{li2021align}, we randomly select $N_v$ frames from the video, with parameters $\frac{N_v}{2}$ frames from the first and second half of the video, respectively. 
We use RandomAugment~\citep{DBLP:conf/nips/CubukZS020} on the frames sampled from each video. 
For all experiments, we use the same random seed (e.g., 42) to ensure reproduction.

\begin{table}[h]
\centering
\scalebox{0.8}{
    \begin{tabular}{lcc}
    \midrule [\heavyrulewidth]
    \textbf{Config}& \textbf{MSRVTT}& \textbf{DiDeMo}\\
    
    \midrule [\heavyrulewidth]
    Optimizer & AdamW &AdamW     \\ 
    Init learning rate (lr)& 2.5e-5 &4e-5\\
    Scaling learning rate & 1.25$\times$lr&1.25$\times$lr\\
    Weight decay & 1e-3 &1e-3     \\ 
    Optimizer momentum & $\beta_1, \beta_2$=0.9,0.98 &$\beta_1, \beta_2$=0.9,0.98     \\ 
    Lr schedule &linear decay&linear decay\\
    Warmup ratio &0.1&0.1\\
    Batch size &64&96\\
    Init $\tau_c$ &0.07 &0.07\\
    Text length &40&50\\
    Frame number &8&8\\
    Training epochs &5&10\\
    Augmentation &RandAug(2,5)&RandAug(2,5)\\
    \midrule [\heavyrulewidth]
    \end{tabular}
    }
\caption{Experimental setup for \textbf{MSRVTT} (both 7k and 9k splits) and \textbf{DiDeMo} text-video retrieval.}
\label{tab:fine-tune_retrieval}
\end{table}
\begin{table}[t]
\centering
\scalebox{0.8}{
    \begin{tabular}{lcc}
    \midrule [\heavyrulewidth]
    \textbf{Config}& \textbf{MSRVTT-QA}& \textbf{MSVD-QA}\\
    
    \midrule [\heavyrulewidth]
    Optimizer & AdamW &AdamW     \\ 
    Init learning rate (lr)& 5e-5 &5e-5\\
    Scaling learning rate & 1.25$\times$lr&1.0$\times$lr\\
    Weight decay & 1e-3 &1e-3     \\ 
    Optimizer momentum & $\beta_1, \beta_2$=0.9,0.98 &$\beta_1, \beta_2$=0.9,0.98     \\ 
    Lr schedule &linear decay&linear decay\\
    Warmup ratio &0.1&0.1\\
    Batch size &96&96\\
    Init $\tau_c$ &0.07 &0.07\\
    Text length &40&40\\
    Frame number & 16&16\\
    Training epochs &10&15\\
    Augmentation &RandAug(2,5)&RandAug(2,5)\\
    \midrule [\heavyrulewidth]
    \end{tabular}
    }
\caption{Experimental setup  for fientuning on \textbf{MSRVTT-QA} and \textbf{MSVD-QA}.}
\label{tab:fine-tune_qa}
\end{table}

\vspace{-5pt}
\section{Comparison with Previous Work}
\begin{table*}[t]
\centering
\scalebox{0.73}{
    \begin{tabular}{lccccc}
    \midrule [\heavyrulewidth]
    \textbf{Methods}&\textbf{Pooling Method}&\textbf{Video Encoder} &\textbf{Text Encoder} &\textbf{Pre-training Data}\\
    \midrule [\heavyrulewidth]
    HT100M~\citet{howto100m}  & Temporal & 2D/3D CNN & word2vec & HowTo100M  \\
    NoiseEST.~\citep{DBLP:conf/aaai/AmraniBRB21} & Temporal  &2D/3D CNN &word2vec&Howto100M \\ 
    ClipBERT~\citep{DBLP:conf/cvpr/LeiLZGBB021} & Temporal  &2D CNN &BERT&COCO, VG \\
    VideoClip~\citep{DBLP:conf/emnlp/XuG0OAMZF21} & Global (Average) & 3D CNN (S3D)&BERT&HowTo100M \\
    ALPRO~\citep{li2021align}& Temporal&TimeSformer&BERT&CC3M, WebVid2M\\
    CLIP4Clip~\citep{DBLP:journals/corr/abs-2104-08860}  & Temporal &ViT (CLIP) &BERT (CLIP) &HowTo100M \\
    BLIP~\citep{blip}& \ Original &ViT&BERT&COCO,VG,CC3M,CC12M,SBU\\
    Frozen~\citep{DBLP:conf/iccv/BainNVZ21}& Global ($\texttt{[CLS]}$) & TimeSformer& DistilBERT  &CC3M,WebVid2M\\
    BridgeFormer~\citep{DBLP:journals/corr/abs-2201-04850} & Original &ViT&DistilBERT&CC3M,WebVid2M \\
    VQA-T~\citep{DBLP:conf/iccv/YangMSLS21}& Global ($\texttt{[CLS]}$)  &3D CNN&DistilBERT&HowToVQA69M\\
    \midrule
    LiteVL (Ours) & Spatial + Temporal + Original & TimeSformer &BERT & -\\

    \midrule [\heavyrulewidth]
    \end{tabular}
    }
\caption{Comparison between previous works with ours in terms of pooling method, video encoder, text encoder and pre-training dataset. 
The pooling method refers to the way to aggregate the video feature based on the 
original video feature $\V_L\in \mathbb{R}^{ (1+ST) \times D}$ (Original), 
before the alignment with text feature. ``Spatial'' and ``Temporal'' denote the  spatially-pooled feature $\V_{f_t} \in \mathbb{R}^{T\times D}$, and temporally-pooled $\V_{f_s}\in \mathbb{R}^{S\times D}$ from $\V_{L}$.}
\label{tab:comparison_method}
\end{table*}

To align video and text features, previous approaches can be generally divided into two categories. On the one hand, common dual-stream models CLIP4Clip \citep{DBLP:journals/corr/abs-2104-08860} fuse \textit{global} video feature from global mean pooling or the $\texttt{[CLS]}$ token, and then interact fused video feature with text feature based on a simple multilayer perceptron head on the top. On the other hand, the cross-attention module is adopted where the key/value are obtained from the aggregated video feature, and the query is obtained from the text feature. Previous methods mainly use two ways to aggregate the original output video features $\V_L$ into $\V_f$ in the video-grounded text encoder:
(i) 
keep the \textit{original}  
features without modification;
(ii) apply mean pooling over the \textit{spatial} or \textit{temporal} dimension. 

We provide a more detailed comparison with related works in Table \ref{tab:comparison_method}.
We list how previous works extract the video features used for the alignment with text features. 
In addition, the video encoder, text encoder, and pre-training data used by differnt methods are also provided.

\section{More Qualitative Results\label{appendix_visualization}}
\begin{figure}[!htbp]
    \centering
    \includegraphics[width=1\linewidth]{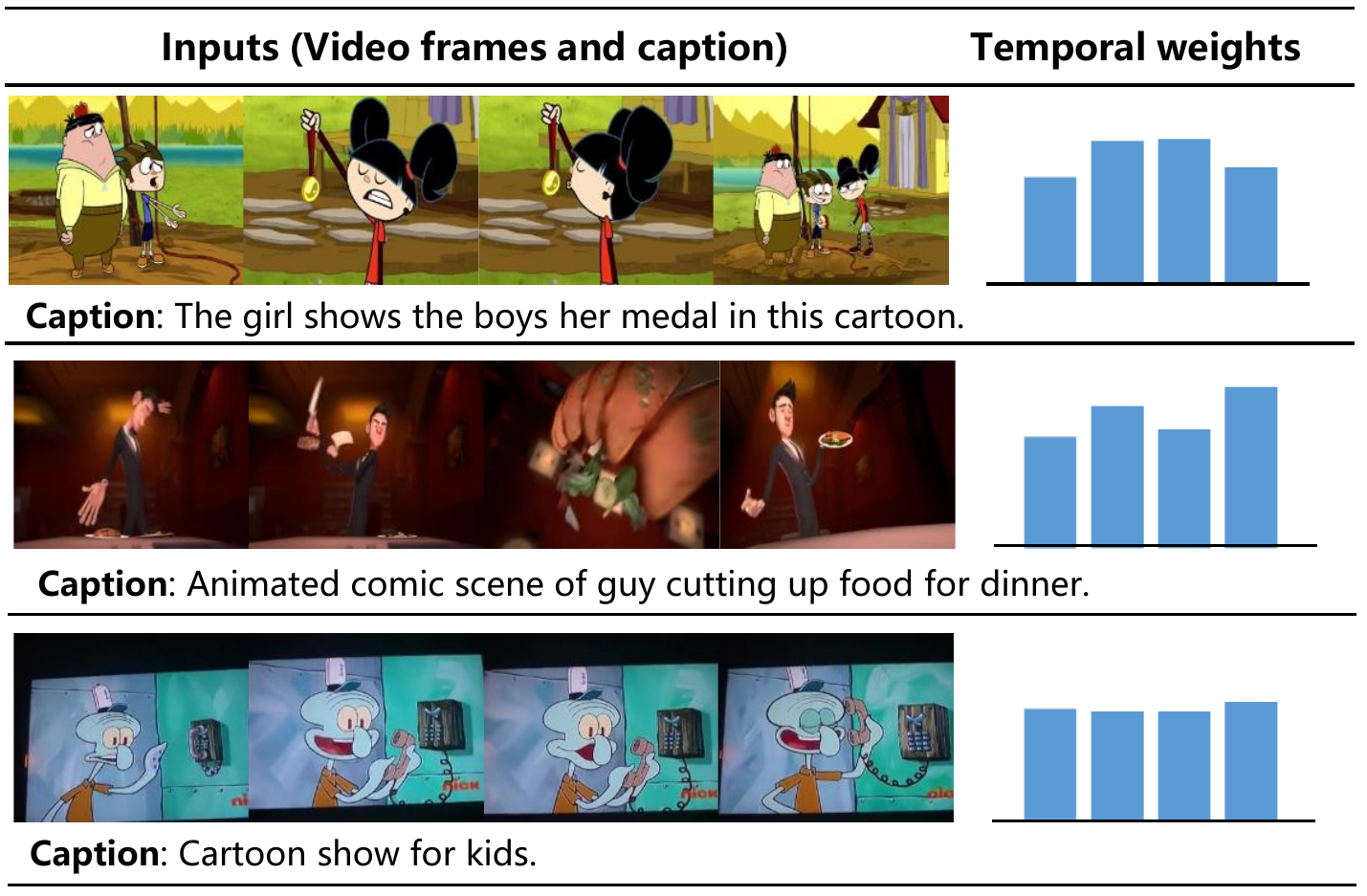}
    \caption{
    Bar plot of temporal weights learned by the text-dependent pooling. 
    }
	\label{fig:visual_appendix}
\end{figure}

We provide more visualizations of temporal weights $\bf{g}_t$  in Figure~\ref{fig:visual_appendix}. 
To better understand how text-dependent pooling affects the decision, we take a closer look at when the proposed text-dependent pooling changes the decision over vanilla pooling (Remark 1). We find that the temporal weights of the changed decisions have a clearly higher standard deviation than the unchanged ones, indicating that text-dependent pooling tends to change the decisions when the different frames are more dissimilar. For instance, for the first case in Figure~\ref{fig:visual_appendix}, its caption \texttt{``The girl shows the boys her medal in this cartoon''} is mainly related to the middle two frames. By assigning higher importance to these two frames, the proposed text-dependent pooling makes a correct decision while the vanilla pooling fails.

\label{appedx:ablation_speed}

\end{document}